\documentclass{article}

\usepackage[colorlinks,
            linkcolor=red,       
            anchorcolor=red,  
            citecolor=green,        
            ]{hyperref}
\usepackage{microtype}
\usepackage{graphicx}
\usepackage{booktabs} 

\usepackage[accepted]{icml2023}

\usepackage{amsmath}
\usepackage{amssymb}
\usepackage{mathtools}
\usepackage{amsthm}

\usepackage[capitalize,noabbrev]{cleveref}

\theoremstyle{plain}
\newtheorem{theorem}{Theorem}[section]

\theoremstyle{definition}
\newtheorem{definition}[theorem]{Definition}

\theoremstyle{remark}

\usepackage[textsize=tiny]{todonotes}

\usepackage{newfloat}
\usepackage{listings}
\usepackage{multirow}
\usepackage{bbm}
\usepackage{amsfonts,amssymb}
\usepackage{enumitem}
\usepackage[T1]{fontenc}
\usepackage{cleveref}
\usepackage{cancel}

\usepackage{wrapfig}
\usepackage{appendix}
\usepackage{units}
\usepackage{algorithm}
\usepackage{algorithmic}
\usepackage{fontawesome5}

\icmltitlerunning{Revisiting Weighted Aggregation in Federated Learning with Neural Networks}

\begin{document}
\setlength{\parskip}{0.3pt plus1.4pt minus0.1pt}

\twocolumn[
\icmltitle{Revisiting Weighted Aggregation in Federated Learning with Neural Networks}





\begin{icmlauthorlist}
\icmlauthor{Zexi Li}{zju}
\icmlauthor{Tao Lin}{wku}
\icmlauthor{Xinyi Shang}{xmu}
\icmlauthor{Chao Wu}{zju}
\end{icmlauthorlist}

\icmlaffiliation{zju}{Zhejiang University, China. Zexi Li <zexi.li@zju.edu.cn>.}
\icmlaffiliation{wku}{Research Center for Industries of the Future, Westlake University, China.}
\icmlaffiliation{xmu}{Xiamen University, China. Xinyi Shang <shangxinyi@stu.xmu.edu.cn>. Work was done during Xinyi's visit to Westlake University}

\icmlcorrespondingauthor{Chao Wu}{chao.wu@zju.edu.cn}
\icmlcorrespondingauthor{Tao Lin}{lintao@westlake.edu.cn}

\icmlkeywords{Federated learning, Deep learning, Weighted aggregation, Neural networks, Training dynamics}

\vskip 0.3in]



\printAffiliationsAndNotice{}  

\begin{abstract}
In federated learning (FL), weighted aggregation of local models is conducted to generate a global model, and the aggregation weights are normalized (the sum of weights is 1) and proportional to the local data sizes. In this paper, we revisit the weighted aggregation process and gain new insights into the training dynamics of FL. First, we find that the sum of weights can be smaller than 1, causing \emph{global weight shrinking} effect (analogous to weight decay) and improving generalization. We explore how the optimal shrinking factor is affected by clients' data heterogeneity and local epochs. Second, we dive into the relative aggregation weights among clients to depict the clients' importance. We develop \emph{client coherence} to study the learning dynamics and find a critical point that exists. Before entering the critical point, more coherent clients play more essential roles in generalization. 
Based on the above insights, we propose an effective method for \textbf{Fed}erated Learning with \textbf{L}earnable \textbf{A}ggregation \textbf{W}eights, named as \textsc{\textbf{FedLAW}} (\href{https://github.com/ZexiLee/ICML-2023-FedLAW}{\faGithub~source code}). Extensive experiments verify that our method can improve the generalization of the global model by a large margin on different datasets and models.
\looseness=-1
\end{abstract}

\section{Introduction} \label{sect:intro}
\begin{figure}[!t]
    \centering
    \includegraphics[width=0.7\columnwidth]{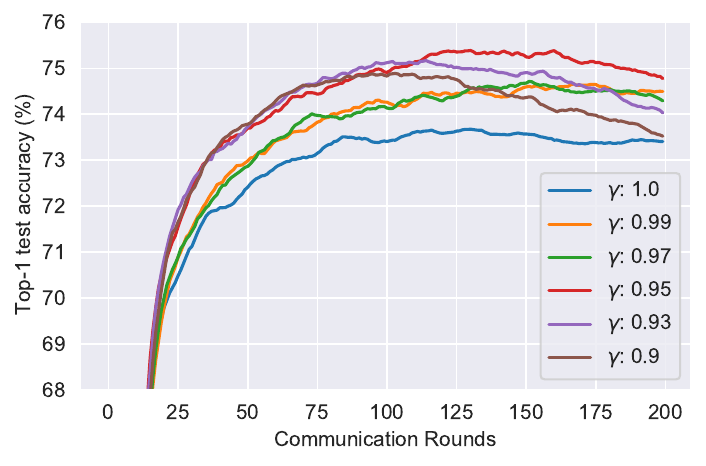}
    \vspace{-1em}
    \caption{\small Test accuracy curves with different $l_1$ norms of aggregation weights ($\gamma$). CIFAR-10 with 20 clients, AlexNet.}
    \label{fig:fix_gamma_testacc}
     \vspace{-0.5cm}
\end{figure}

Federated learning (FL) \citep{mcmahan2017communication,DBLP:journals/spm/LiSTS20,wang2021field,lin2020ensemble,li2022mining} is a promising distributed optimization paradigm where clients' data are kept local, and a central server aggregates clients' local gradients for collaborative training. 
In FL, weighted aggregation of local models is conducted to generate a global model. In FL, when aggregating local models, it is a common practice that the aggregation weights should be normalized (the sum of weights, i.e.\ the $l_1$ norm, notated as $\gamma$, is equal to 1) and proportional to the local data sizes. However, due to the non-convexity \citep{allen2019convergence,li2018visualizing}, over-parameterization \citep{allen2019convergence,zou2019improved}, scale invariance \citep{li2018visualizing,dinh2017sharp,kwon2021asam}, and other unique properties of deep neural networks (DNNs), there is a gap between theory and empirical practice when the models are DNNs. An intuitive example is shown in \autoref{fig:fix_gamma_testacc}, we find that smaller $\gamma$ may be beneficial to generalization, which challenges the previous convention in theory that aggregation weights should be normalized as 1. But what is the mechanism behind and what is the optimal $\gamma$ under different FL environments? It requires further investigation.

Thus, in this paper, we revisit and rethink the weighted aggregation process to understand the training dynamics of FL and gain some intriguing insights. 
\begin{center}
    \emph{How can the aggregation weights be assigned to generate a global DNN model with better generalization?}
\end{center}
Towards this question, we find two aspects that matter most: \textbf{(1) the $l_1$ norm of aggregation weights ($\gamma$); }

\textbf{(2) the relative weights within the sampled clients ($\boldsymbol{\lambda}$).} 

To gain insights, we leverage the advantage of the server in FL that we learn the aggregation weights on a global-objective-consistent proxy dataset by gradient descent. The learned weights are the optimal weight candidates at each round and can reflect the training dynamics. 

For (1), we identify the \emph{global weight shrinking} effect in FL when $\gamma$ is smaller than 1, which is analogous to weight decay regularization \citep{loshchilov2018decoupled,lewkowycz2020training,xie2020understanding} in centralized training. However, a small value of $\gamma$---as stated in~\autoref{fig:fix_gamma_testacc}---will cause negative effects; therefore, there exists an optimal $\gamma$ that balances the regularization and optimization. 
We fix $\boldsymbol{\lambda}$ and learn $\gamma$ (cf.\ \autoref{sect:gws}) on the proxy dataset to explore how the optimal shrinking factor (i.e. the $l_1$ norm of $\gamma$) is affected by clients' heterogeneity and local epochs. 

For (2), we study how $\boldsymbol{\lambda}$ should be assigned to the local models to obtain a more generalized global model and how $\boldsymbol{\lambda}$ can reflect clients' importance in training dynamics. We fix $\gamma$ and learn $\boldsymbol{\lambda}$ (cf.\ \autoref{sect:hyperplane}) on the proxy dataset to study \emph{client coherence} that includes: (i) \emph{local gradient coherence}, the importance of clients in different learning periods; (ii) \emph{heterogeneity coherence}, the consistency between the sum objective of sampled clients and the global objective.

Based on the insights, we propose an effective method for \textbf{Fed}erated Learning with \textbf{L}earnable \textbf{A}ggregation \textbf{W}eights, named as \textsc{\textbf{FedLAW}} (cf.\ \autoref{sect:fedawo}). Extensive experiments verify that our method can improve the generalization of the global model by a large margin on different datasets and models. Moreover, it is validated that \textsc{FedLAW} is still robust when the proxy dataset is small or shifted from the global distribution and corrupted clients exist.

\textbf{Specifically, our contributions are two-folded.}
\begin{itemize}[leftmargin=*,nosep]
    \item As our main contribution, we revisit and rethink the weighted aggregation in FL with DNNs and identify some interesting findings (see below take-aways). Especially, we find that smaller $l_1$ norms of aggregation weights may be beneficial to generalization, which challenges the previous normalized convention. This is also the first paper that introduces global regularization in FL, and we explore how to adaptively control such regularization.
    \looseness=-1
    \item We showcase the applicability of these insights, and devise a simple yet effective method \textsc{FedLAW}, which largely boosts the generalization of global models. The effectiveness and robustness of \textsc{FedLAW} are validated by extensive experiments.
\end{itemize}

\textbf{We summarize our key take-away messages of the understandings as follows.}
\begin{itemize}[nosep,leftmargin=*]
    \item \textit{Global weight shrinking} regularization effectively improves the generalization performance. 
    \looseness=-1
    \begin{itemize}[leftmargin=12pt, nosep]
    \item The magnitude of the global gradient (i.e.\ uniform average of local updates) determines the optimal weight shrinking factor. 
    A larger norm of the global gradient requires stronger regularization, in the cases when (i) the number of local epochs is larger; (ii) the clients' data are more IID; (iii) during training before the global model is near convergence.
    \looseness=-1
    \item The effectiveness of global weight shrinking is stemmed from flatter loss landscapes of the global model as well as the improved local gradient coherence after the critical point.\footnote{
    Different from the latter observations (w/o affecting the training dynamics), applying global weight shrinking results in a positive local gradient coherence after the critical point and the learning can benefit from it.
    \looseness=-1
    }
    \end{itemize}
    \item Our novel concept of \textit{client coherence} depicts the training dynamics of FL, from the aspects of \emph{local gradient coherence} and \emph{heterogeneity coherence}.
    \begin{itemize}[leftmargin=12pt, nosep]
    \item Local gradient coherence refers to the averaged cosine similarities of clients' local gradients. 
    A critical point (from positive to negative) exists in the curves of local gradient coherence during the training. Generalization can benefit when the local gradient coherence is positive and more dominant.
    \looseness=-1
    \item Heterogeneity coherence refers to the distribution consistency between the global data and the sampled one (i.e.\ data distribution of a cohort of sampled clients) in each round. 
    Increasing the heterogeneity coherence by reweighting the sampled clients could also improve the training performance.
    \looseness=-1
    \end{itemize}
\end{itemize}

\section{Related Works} \label{sect:related_works}
\textbf{Model aggregation in FL. }
There are previous works that try to learn the aggregation weights on given datasets by gradient descent. \textsc{Auto-FedAvg} \citep{xia2021auto} learns aggregation weights on different institutional medical data to realize personalized medicine, while \textsc{L2C} matches similar peers in decentralized FL \citep{li2022learning} by learning aggregation weights on local datasets. These works all adopt the normalized aggregation weights ($\gamma$=1) without discovering the global weight shrinking effect, and they focus on \textit{personalization} while we focus on \textit{generalization}. Besides, they fail to understand the FL's dynamics from the learned weights for further insights, e.g., identifying the significance of client coherence. 
Ensemble distillation methods are used to improve the generalization of global models after weighted aggregation. \textsc{FedDF} \citep{lin2020ensemble} uses the local models as teachers and finetune the global model via ensemble distillation; while in \textsc{FedBE} \citep{DBLP:conf/iclr/ChenC21}, Bayesian ensemble distillation is further introduced.  Since they also require a proxy dataset on the server, we will compare them with our proposed \textsc{FedLAW} in \autoref{sect:fedawo}. Additionally, server-side stochastic weight averaging and client-side sharpness-aware minimization are incorporated to make the global model converge to a flatter minimum \citep{caldarola2022improving}; distributionally robust optimization is also introduced to realize more robust federated averaging \citep{deng2020distributionally,wu2022drflm}; but these works are orthogonal to our paper.

\textbf{Training dynamics of DNNs in centralized learning. }
Our insights into global weight shrinking and client coherence in FL are analogous to weight decay and gradient coherence in centralized learning. \textit{Weight decay: }The optimal weight decay factor is approximately inverse to the number of epochs, and the importance of applying weight decay diminishes when the training epochs are relatively long \citep{loshchilov2018decoupled,lewkowycz2020training,xie2020understanding}. The effectiveness of weight decay may be explained by the caused (i) larger effective learning rate \citep{zhang2018three,wan2021spherical}, and (ii) flatter loss landscape \citep{lyu2022understanding}. \textit{Gradient coherence:} Gradient coherence, or sample coherence, is a crucial technique for understanding the training dynamics of mini-batch SGD in centralized learning \citep{chatterjee2019coherent,zielinski2020weak,chatterjee2020making,fort2019stiffness}.
The gradient coherence measures the pair-wise gradient similarity among samples. If they are highly similar, the overall gradient within a mini-batch will be stronger in certain directions, resulting in a dominantly faster loss reduction and better generalization \citep{chatterjee2019coherent,zielinski2020weak,chatterjee2020making}. The critical period exists in mini-batch SGD, captured by the gradient coherence: the low coherence in the early training phase damages the final generalization performance, no matter the value of coherence controlled later \citep{chatterjee2020making}. 
In \autoref{sect:gws} and \autoref{sect:hyperplane}, we will show similar findings can be drawn in FL's dynamics, and some new insights are discovered. 

Due to space limits, the detailed discussions about related works can be found in \autoref{appendix:related_works}.

\section{Preliminary and Problem Setup} \label{sect:preliminary}
FL usually involves a server and $n$ clients to jointly learn a global model without data sharing, which is originally proposed in \citep{mcmahan2017communication}. 
Denote the set of clients by $\mathcal{S}$, the local dataset of client $i$ by $\mathcal{D}_i=\{(x_j, y_j)\}_{j=1}^{N_i}$, the sum of clients' data by $\mathcal{D} = \bigcup_{i\in\mathcal{S}} \mathcal{D}_i$. The IID data distributions of clients refer to each client's distribution $\mathcal{D}_i$ is IID sampled from $\mathcal{D}$. However, in practical FL scenarios, \textit{heterogeneity} exists among clients that their data are \textit{NonIID} with each other. In this paper, we use Dirichlet sampling, which is widely used in FL literature \citep{lin2020ensemble,li2020federated,acar2020federated}, to synthesize \textit{\textbf{client heterogeneity} (controlled by \textbf{$\alpha$}, the smaller, the more NonIID)}. During FL training, clients iteratively conduct local training and communicate with the server for model updating. In the local training, \textit{\textbf{the number of local epochs} is \textbf{$E$}}; when $E$ is larger, the communication is more efficient but the updates are more asynchronous. Since $\alpha$ and $E$ are the key factors affecting FL's training, in this paper, we study how $\alpha$ and $E$ affect the training dynamics of FL from the perspective of weighted aggregation.

Denote the global model and the client $i$'s local model in communication round $t$ by $\mathbf{w}_{g}^{t}$ and $\mathbf{w}_{i}^{t}$. 
In each round, clients' local models are initialized as the global model that $\mathbf{w}_{i}^{t} \leftarrow \mathbf{w}_{g}^{t}$, and clients conduct local training in parallel. In each local training epoch, clients conduct SGD update with a local learning rate $\eta_l$, and each SGD iteration shows as
\begin{align}
\mathbf{w}_{i}^{t} \leftarrow \mathbf{w}_{i}^{t} - \eta_l \nabla \ell(B_k, \mathbf{w}_{i}^{t}), \text{ for }  k=1,2,\cdots,K, \label{eqt_client}
\end{align}
where $\ell$ is the loss function and $B_k$ is the mini-batch sampled from $\mathcal{D}_i$ at the $k$-th iteration. 
After the client local updates, the server samples $m$ clients for aggregation. The client $i$'s pseudo gradient of local updates is denoted as $\textbf{g}_{i}^{t} = \mathbf{w}_{g}^{t} - \mathbf{w}_{i}^{t}$. 
Then, the server conducts weighted aggregation to merge the local models (or the pseudo gradients) into a new global model\footnote{As in \autoref{eqt_FedAvg}, FL's aggregation can be formulated into the aggregation of clients' local models (left) or clients' pseudo gradients (right). The two kinds of the formulation are equal, while we adopt the aggregation of models here for brevity.}. 
\begin{small}
\begin{align}
\mathbf{w}_{g}^{t+1} = \sum_{i=1}^{m} \mu_{i}\textbf{w}_{i}^{t} = \Vert \boldsymbol{\mu} \Vert_{1}\textbf{w}_{g}^{t} -\eta_g\sum_{i=1}^{m} \mu_{i}\textbf{g}_{i}^{t}, \text{ s.t. } \mu_{i} \geq 0,
\label{eqt_FedAvg}
\end{align}
\end{small}
where $\boldsymbol{\mu} = [\mu_1, \dots, \mu_m]$ is the aggregation weights, $\eta_g=1$ is the global learning rate. For vanilla \textsc{FedAvg}, it adopts a normalized weights proportional to the data sizes, $\mu_{i}=\frac{|\mathcal{D}_{i}|}{|\mathcal{D}|}, \mathcal{D} = \bigcup_{i\in\mathcal{S}} \mathcal{D}_i$. In this paper, we assume the aggregation weights are not normalized which means the $l_1$ norm is not necessarily equal to 1. We study the effects of the $l_1$ norm and relative weights independently by decouple $\boldsymbol{\mu}$ into $\{\gamma, \boldsymbol{\lambda}\}$, which satisfies $\gamma = \Vert \boldsymbol{\mu} \Vert_{1}, \lambda_i = \frac{\mu_i}{\Vert \boldsymbol{\mu} \Vert_{1}}$. Thus, \autoref{eqt_FedAvg} can be reformulated into
\begin{align}
\mathbf{w}_{g}^{t+1} = \gamma\sum_{i=1}^{m} \lambda_{i}\textbf{w}_{i}^{t}, \text{ s.t. } \gamma > 0, \lambda_{i} \geq 0, \Vert \boldsymbol{\lambda} \Vert_{1}=1.
\label{eqt_reformulate_fedavg}
\end{align}
Vanilla \textsc{FedAvg} is a special case where $\gamma=1, \lambda_{i}=\frac{|\mathcal{D}_{i}|}{|\mathcal{D}|}, \forall i \in [m]$. When $\gamma<1$, it will cause weight shrinking of the global model, so in this case, we also call $\gamma$ the shrinking factor.

\textbf{Clarification on the proxy dataset.} We study global weight shrinking\footnote{We use the word ``shrink'' instead of ``decay'' as it shrinks the global model rather than decaying the model by subtracting a decay term (used in traditional weight decay). Similar ``shrink'' can be found in \cite{li2020understanding}.\looseness=-1} ($\gamma$) in \autoref{sect:gws} and client coherence ($\boldsymbol{\lambda}$) in \autoref{sect:hyperplane} through respectively learning $\gamma$ and $\boldsymbol{\lambda}$ while fixing another on a server proxy dataset. 
The considered proxy dataset has the same distribution as the global learning objective (i.e.\ a class-balanced case in this paper; e.g.\ 2000 balanced samples in CIFAR-10), thus the learned aggregation weights $\{\gamma, \boldsymbol{\lambda}\}$ can reflect the contributions of clients and the optimal regularization factor towards this global objective. We note that this case of the proxy dataset is for understanding only, and we will validate the effectiveness and robustness of our proposed \textsc{FedLAW} on tiny (e.g.\ 100 samples in CIFAR-10) or biased (e.g.\ long-tailed) proxy datasets. 
For concision, in \autoref{sect:gws} and \autoref{sect:hyperplane}, if not mentioned otherwise, we all use CIFAR-10 as the dataset and SimpleCNN as the model. Experiments on more datasets and models are shown in \autoref{sect:fedawo} and Appendix.

\section{Global Weight Shrinking} \label{sect:gws}
\subsection{Global Weight Shrinking and Its Impacts on Optimization} \label{subsect:gws1}

Setting $\gamma < 1$ results in the global weight shrinking regularization. \autoref{table:general_GWS} and \autoref{fig:fix_gamma_testacc} report the results on CIFAR-10 with different $\gamma$. 
It can be observed that the \emph{global weight shrinking may improve generalization, depending on the choice of $\gamma$.} The smaller $\gamma$, the stronger regularization effect. Given a setting, there exists an optimal $\gamma$ that balances the regularization and optimization, and deviation from this value, whether smaller or larger, may result in inferior performance. 
More results about the fixed $\gamma$ can be found in \autoref{appdx_table:general_GWS} in Appendix.

\begin{figure}[!t]\centering
    \vspace{-0.25cm}
    \centering
    \includegraphics[width=0.48\columnwidth]{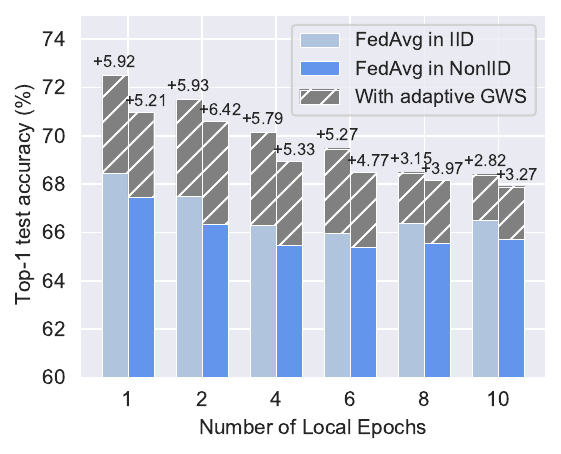}
    \includegraphics[width=0.48\columnwidth]{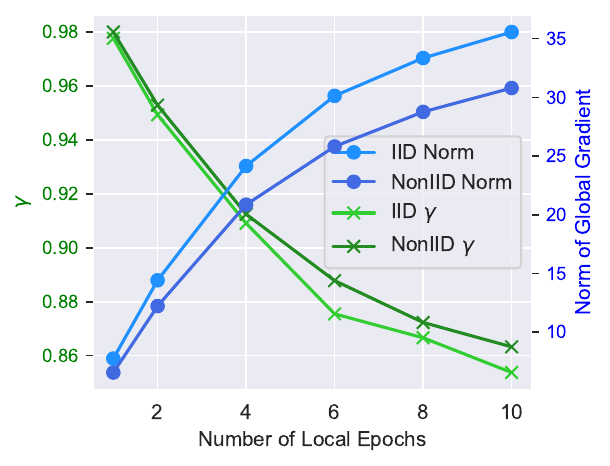}
    \vspace{-1em}
    \caption{\small \textbf{Left:} Test accuracy gains of adaptive GWS. \textbf{Right:} The optimal $\gamma$ and the norm of global gradient. $\alpha \in \{100,1\}$.}
    \label{fig:fix_gamma_test_gain}
     \vspace{-0.3cm}
\end{figure}

\begin{table}[t]
\centering
\vspace{-0.1cm}
\caption{\small Impact of fixed $\gamma$ across different architectures in both IID ($\alpha=100$) and NonIID ($\alpha=1$) settings ($E=2$).}
\resizebox{0.48\textwidth}{!}{
\begin{tabular}{c|c|cccccc@{}}
\toprule  
&$\gamma$ &1.0 &0.99 &0.97 &0.95 &0.93&0.9 \\ 
\midrule
\multirow{3}*{IID}&SimpleCNN & 65.53 &67.60 &69.20 &69.52 &\textbf{70.16} &69.83 \\
&AlexNet&74.16 &74.80 &\textbf{75.54} &75.24 &75.25 &75.03 \\
&ResNet8&75.51 &76.64 &76.80 &\textbf{77.87} &76.80 &76.74 \\
\midrule
\multirow{3}*{NonIID}&SimpleCNN &65.58 &67.04 &68.36 &68.66 &\textbf{69.28} &68.93
\\
&AlexNet&73.56 &73.83 &74.37 &\textbf{74.45} &74.40 &74.24
\\
&ResNet8&75.02 &76.06 &75.73 &\textbf{77.00} &75.04 &75.31 \\
 \bottomrule
\end{tabular}
}
\label{table:general_GWS}
\vspace{-0.4cm}
\end{table}

\subsection{Adaptive Global Weight Shrinking and Training Dynamics} \label{sect:opt_GWS_by_GD}
We discover how to set an appropriate $\gamma$ to balance regularization and optimization. We first expand the right of \autoref{eqt_FedAvg} as follows.

\vspace{-0.3cm}
\begin{small}
\begin{equation} \label{equ:gws_update}
    \mathbf{w}_{g}^{t+1} = \gamma(\mathbf{w}_{g}^{t} -\eta_{g}\textbf{g}_{g}^{t})
    = \mathbf{w}_{g}^{t} -\gamma\eta_{g}\textbf{g}_{g}^{t} -(1-\gamma)\mathbf{w}_{g}^{t}.
\end{equation}
\end{small}
We refer $(1-\gamma)\mathbf{w}_{g}^{t}$ as the pseudo gradient of global weight shrinking (\textit{regularization term}) and $\gamma\eta_{g}\textbf{g}_{g}^{t}$ is the global averaged gradient (\textit{optimization term}). We reckon that a larger optimization term requires a larger regularization term, which means the magnitude of the global pseudo gradient $\textbf{g}_{g}^{t}$ determines the optimal shrinking factor $\gamma$ in the way that larger $\textbf{g}_{g}^{t}$, smaller $\gamma$ (stronger regularization). 

To verify our hypothesis, we achieve adaptive global weight shrinking (adaptive GWS) on the proxy dataset, which learns an optimal $\gamma$. Adaptive GWS adopts the update in \autoref{eqt_reformulate_fedavg} and uses $\{\gamma = \gamma^{*}, \lambda_{i}=\frac{|\mathcal{D}_{i}|}{|\mathcal{D}|}\}$ where
\begin{equation} \label{equ:opt_gamma}
\gamma^{*} = \mathop{\arg\min}\limits_{\gamma}  \mathcal{L}_{proxy} (\gamma\cdot\sum_{i=1}^{m} \frac{|\mathcal{D}_{i}|}{|\mathcal{D}|}\textbf{w}_{i}^{t}), 
 \text{ s.t. } \gamma > 0.
\end{equation}
Adaptive GWS largely improves the generalization. 
From the left of \autoref{fig:fix_gamma_test_gain}, adaptive GWS can improve the performance of \textsc{FedAvg} by a large margin in both IID and NonIID settings. Furthermore, adaptive GWS is more beneficial when the number of local epochs is small. 

\begin{figure}[t]
    \centering
     \vspace{-0.25cm}
    \includegraphics[width=0.49\columnwidth]{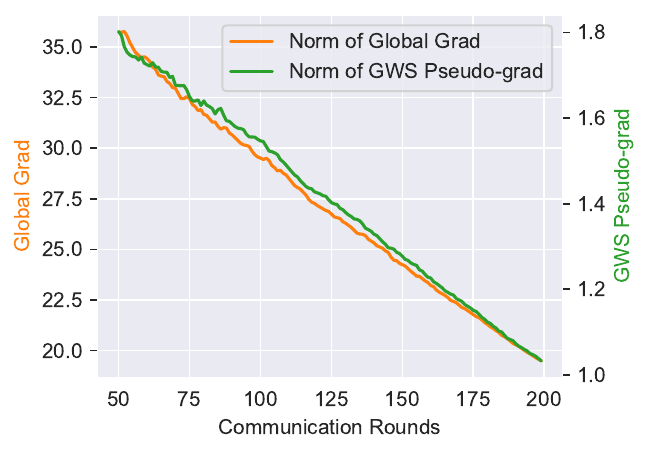}
    \includegraphics[width=0.49\columnwidth]{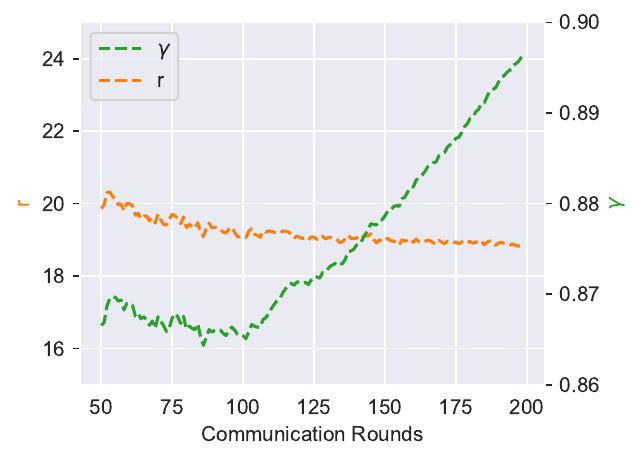}
    \vspace{-1em}
    \caption{\small \textbf{Left:} Norm of two gradients in adaptive GWS. \textbf{Right:} The optimal $\gamma$ and $r$ in adaptive GWS, where $r$ is the ratio of the global gradient and the regularization pseudo gradient.
    }
    \label{fig:gws_r}
     \vspace{-0.47cm}
\end{figure}

\textbf{1) Understanding the balance between optimization and regularization.} Further, through the learned optimal $\gamma$, we verify the balance between optimization and regularization from the right of \autoref{fig:fix_gamma_test_gain} and \autoref{fig:gws_r}. A larger norm of the global gradient requires stronger regularization, in the cases when (i) the number of local epochs is larger; (ii) the clients' data are more IID; (iii) during training before the global model is near convergence (on the contrary, when the model is near convergence, smaller regularization is needed).
\begin{itemize}[leftmargin=*,nosep]
    \item As shown in the right blue Y-axis of right \autoref{fig:fix_gamma_test_gain}, the norm of global gradient $\Vert \gamma\eta_{g}\textbf{g}_{g}^{t} \Vert$ increases when the number of local epochs increases and data become IID. As a result, the optimal value of $\gamma$ (shown in the left green Y-axis) becomes smaller in order to produce a larger weight shrinking pseudo gradient $\Vert (1-\gamma)\mathbf{w}_{g}^{t} \Vert$ to regularize the optimization. More results regarding how heterogeneity affects the optimal $\gamma$ can be found in \autoref{appdx_fig:gws_testacc_gamma} in Appendix.
    \looseness=-1
    \item In \autoref{fig:fix_gamma_testacc}, GWS with smaller fixed $\gamma$ will cause performance degradation in the late training. This is due to the conflicts of decaying global pseudo gradient and non-decaying regularization pseudo gradient.  
    In \autoref{fig:gws_r}, while the norm of the global gradient is decaying, adaptive GWS learns a rising optimal $\gamma$ to keep the GWS pseudo gradient decay proportionally. 
    As a result, the ratio of two gradient terms remains steady at around 19 to maintain the balance between optimization and regularization.
\end{itemize}

\begin{figure*}[t]
    \centering
     \includegraphics[width=0.6\columnwidth]{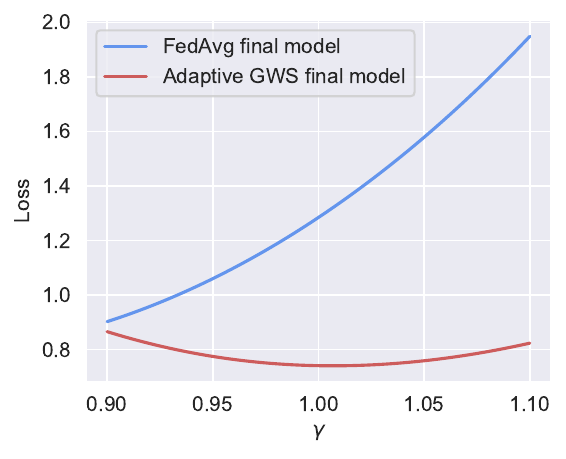}
    \includegraphics[width=0.6\columnwidth]{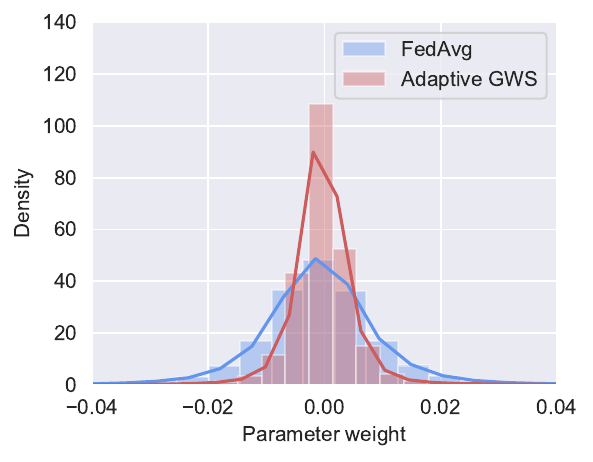}
    \includegraphics[width=0.6\columnwidth]{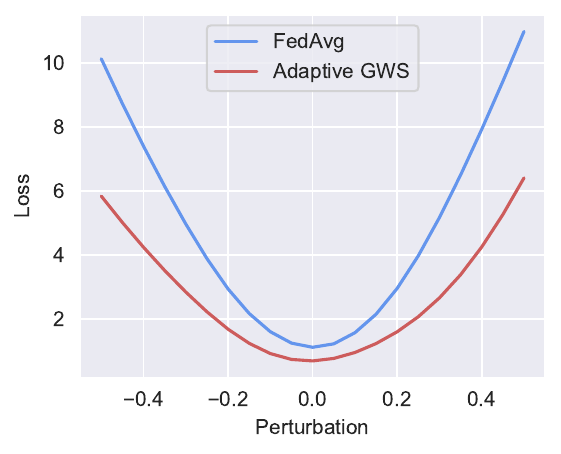}
    \vspace{-1em}
    \caption{\small \textbf{General understanding of adaptive GWS.} \textbf{Left:} Scale invariance property of DNNs indicates that if the network is rescaled by $\gamma$, the function of the model remains similar. \textbf{Middle:} The histogram of final models' parameters shows that adaptive GWS makes more model parameters close to zero, nearly twice as many as \textsc{FedAvg}. \textbf{Right:} The loss landscape is perturbed based on the Top-1 Hessian eigenvector of the final models, which shows that the model with adaptive GWS has flatter curvature and smaller loss.} 
    \label{fig:general_understand}
\end{figure*}

\textbf{2) The mechanisms behind adaptive GWS.} We provide an in-depth general understanding of how adaptive GWS works and  why it can improve generalization. \looseness=-1
\begin{itemize}[nosep,leftmargin=*]
    \item \textbf{General understanding.} 
    \begin{itemize}[leftmargin=12pt, nosep]
        \item \textbf{Scale invariance.} Adaptive GWS learns a dynamic shrinking factor $\gamma$ in each round to shrink the global model's parameter. The method is effective due to the scale invariance property of DNNs \citep{li2018visualizing,dinh2017sharp,kwon2021asam}, which states that the function of a DNN remains similar or the same even when a factor rescales the model weights due to the non-linearity of activation functions or the normalization layer in DNNs. We show an intuitive understanding of scale invariance on the left figure of \autoref{fig:general_understand}, where the final models are rescaled by $\gamma$, and the loss function of the adaptive GWS's final model remains similar while the \textsc{FedAvg}'s final model even has a smaller loss when $\gamma < 1$.
        \item \textbf{Small model parameters.} The shrinking effect in each round can result in smaller model parameters of final global models, which is similar to weight decay. 
        The parameter weight histogram is demonstrated in the middle figure of \autoref{fig:general_understand}. The final model of adaptive GWS has more model parameters close to zero, nearly twice as many as \textsc{FedAvg}. 
        
    \end{itemize}
    \item \textbf{Why adaptive GWS can improve generalization.}
    \begin{itemize}[leftmargin=12pt, nosep]
        \item \textbf{Flatter loss landscapes.} One perspective of explaining the generalization of DNNs is through the flatness of the loss landscape. Previous works have shown that flatter curvature in loss landscape can indicate better generalization \citep{fort2019large,foret2020sharpness,li2018visualizing}. \cite{lyu2022understanding} shows that weight decay of mini-batch SGD can result in flatter landscapes in DNNs with normalization layers. 
        We also observe the similar phenomenon that \emph{adaptive GWS improves generalization by seeking flatter minima in FL}, as shown in the right figure of \autoref{fig:general_understand}. Other metrics of flatness also demonstrate similar results (\autoref{appdx:fig_general_understand} in Appendix).
    \end{itemize}
\end{itemize}

\textbf{3) The relation between adaptive GWS and local weight decay.}
Our proposed adaptive GWS can provide weight regularization from the global perspective, which is analogous to weight decay in mini-batch SGD. Importantly, GWS has a unique sparse regularization frequency that only changes the model weight in each round, resulting in stronger regularization. In GWS, $1 - \gamma$ is near 0.1, whereas the factor of weight decay is often around $10^{-4}$. Notably, the two methods are not conflicted in FL, and we conduct experiments on implementing weight decay in the local SGD solver and global weight shrinking on the server simultaneously. As shown in \autoref{table:local_wd}, \emph{adaptive GWS is compatible with local weight decay and can further improve performance.} Unlike local weight decay, adaptive GWS is hyperparameter-free and effective. It can adaptively set $\gamma$ to maximize the benefit of weight regularization. As the local weight decay becomes stronger, the learned $\gamma$ is larger, resulting in weaker GWS regularization. More analysis about global weight shrinking can be found in \autoref{appendix:analysis_gws} in Appendix.

\textbf{4) Insights from FL's adaptive GWS to mini-batch SGD.} 
FL's adaptive GWS leverages the advantage of the server that learns an adaptive shrinking factor globally. It is promising that similar ideas can be adopted in mini-batch SGD by learning the hyperparameter of weight decay on a small proportion of training data. This may realize hyperparameter-free optimization, and we leave it for future works.

\begin{table}[t]
\centering
\caption{\small	Adaptive GWS with different local weight decay factors ($E=2$). IID ($\alpha=100$), NonIID ($\alpha=1$).}
\resizebox{0.48\textwidth}{!}{
\begin{tabular}{c|c|cccccc@{}}
\toprule  
&\textbf{Local weight decay} & 0    & 5e-5  & 1e-4  & 5e-4  & 1e-3 \\ 
\midrule
\multirow{3}*{IID}&\textsc{FedAvg}                                       & 66.43                           & 66.20                           & 66.45                           & 67.51                           & 68.66\\
&Adaptive GWS                                 & \textbf{71.47} & \textbf{71.36} & \textbf{71.35} & \textbf{71.44} & \textbf{71.54}\\
&$\gamma$ of Adaptive GWS                     & 0.9472                          & 0.9477                          & 0.948                           & 0.9493                          & 0.953\\ 
\midrule
\multirow{3}*{NonIID}&\textsc{FedAvg}                                  & 65.35                           & 65.19                           & 65.77                           & 66.37                           & 67.4                            \\
&Adaptive GWS & \textbf{70.31} & \textbf{69.93} & \textbf{70.44} & \textbf{70.47} & \textbf{69.99} \\
&$\gamma$ of Adaptive GWS   & 0.9492                          & 0.9499                          & 0.9505                          & 0.9529                          & 0.9561                          \\ 
\bottomrule
\end{tabular}
}
\label{table:local_wd}
\vspace{-10pt}
\end{table}

\begin{figure*}[!t]
\centering
\includegraphics[width=0.86\columnwidth]{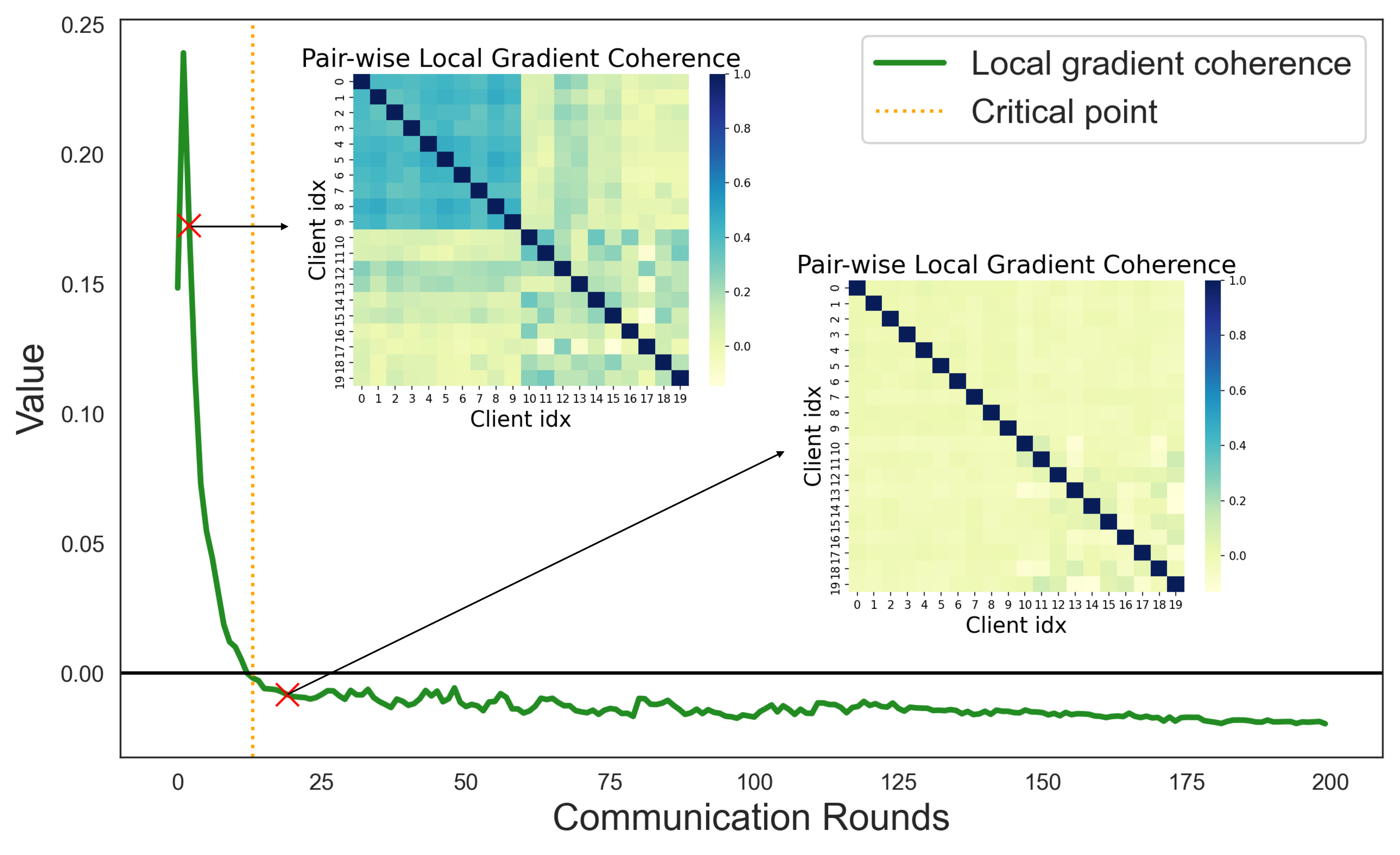}
\includegraphics[width=0.53\columnwidth]{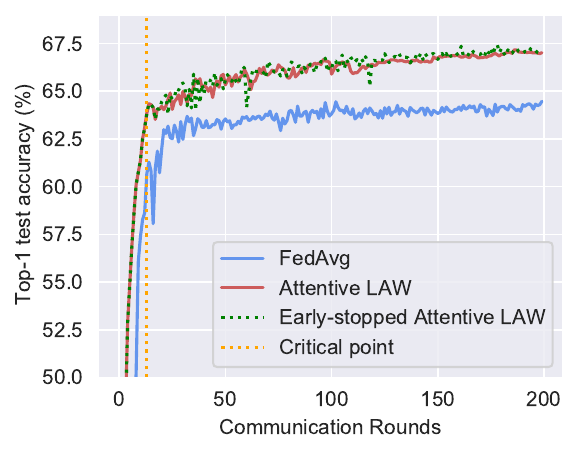}
\includegraphics[width=0.53\columnwidth]{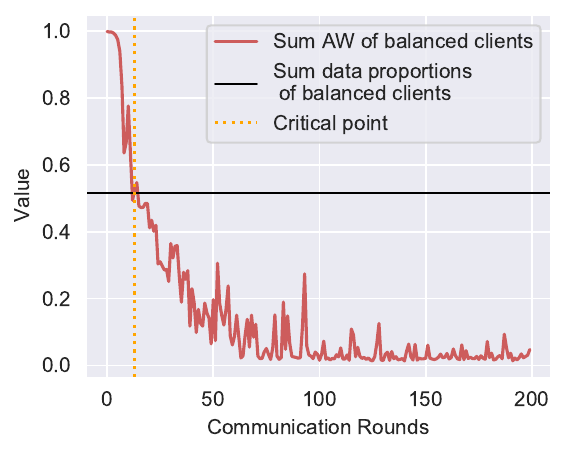}
\vspace{-1em}
\caption{\small \textbf{Training dynamics of attentive LAW in terms of local gradient coherence.} Clients indexed 0-9 have balanced class distributions and 10-19 are imbalanced, $E=3$. \textbf{Left:} Local gradient coherence.  \textbf{Middle and Right:} The performance and aggregation weights of attentive LAW and early-stopped attentive LAW.}
\label{fig:hyperplane_noniid}
\vspace{-0.5cm}
\end{figure*}

\section{Client coherence} \label{sect:hyperplane}
\subsection{Basic Concept and Formulation}
Inspired by gradient coherence in mini-batch SGD \citep{chatterjee2019coherent,zielinski2020weak,chatterjee2020making}, we study \emph{client coherence} in FL through weighted aggregation, which indicates how clients strengthen and complement each other to achieve better generalization.
There are two aspects, the \emph{local gradient coherence} of clients' model updates and the \emph{heterogeneity coherence}.

 \textbf{Local Gradient Coherence.}
The gradient coherence in mini-batch SGD is at the data sample level. In FL, the concept of gradient coherence is extended to the level of clients, where we refer to it as "client coherence". Specifically, we study the similarity of local gradients among clients, as it has been shown that aggregating similar gradients leads to stronger global gradients, thereby improving generalization. We deduce the gradient coherence in mini-batch SGD and local gradient coherence in FL under a unified equation below:\looseness=-1 

\vspace{-0.3cm}
\begin{align} \label{equ:gradient_coherence}
\begin{split}
    \displaystyle
    &\Delta\mathcal{L}^{t} 
    = \mathcal{L}(\mathbf{w}^{t}-\eta\textbf{g}^{t}) - \mathcal{L}(\mathbf{w}^{t}) \approx -\eta \cdot \langle \textbf{g}^{t}, \textbf{g}^{t} \rangle \\
    &= -\eta \cdot \langle \sum_{i=1}^{m}\lambda_{i}\textbf{g}_{i}^{t}, \sum_{i=1}^{m}\lambda_{i}\textbf{g}_{i}^{t} \rangle \\
    &= -\eta \cdot(\sum_{i=1}^{m}\lambda_{i}^{2}\Vert \textbf{g}_{i}^{t}\Vert^{2} + \sum_{i, j, i \neq j}\lambda_{i}\lambda_{j}\langle \textbf{g}_{i}^{t}, \textbf{g}_{j}^{t}\rangle)\\
    &= -\eta \cdot(\sum_{i=1}^{m}\lambda_{i}^{2}\Vert \textbf{g}_{i}^{t}\Vert^{2} + \sum_{i, j, i \neq j}\lambda_{i}\lambda_{j}\cos( \textbf{g}_{i}^{t}, \textbf{g}_{j}^{t})\Vert \textbf{g}_{i}^{t}\Vert\Vert \textbf{g}_{j}^{t}\Vert) \,.
\end{split}
\end{align}

\autoref{equ:gradient_coherence} is a Taylor expansion of the loss function within one update. In mini-batch SGD, $t$ is the iteration step, $m$ is the batch size, and $\textbf{g}_{i}^{t}$ is the gradient of a sample $i$ at iteration $t$. Typically, there is no weighted averaging in a mini-batch, so $\forall i \in [m],\ \lambda_{i} = 1$. In FL, $t$ is the communication round, $\mathbf{w}^{t}$ is the global model on the server at round $t$, $m$ is the cohort size, $\textbf{g}_{i}^{t}$ denotes the local gradient of client $i$ at round $t$, and $\lambda_i$ is the aggregation weight of client $i$. The term $\cos( \textbf{g}_{i}^{t}, \textbf{g}_{j}^{t})$ means the cosine similarity between the gradients of clients $i$ and $j$, defined as $\nicefrac{\langle \textbf{g}_{i}^{t}, \textbf{g}_{j}^{t}\rangle }{\Vert \textbf{g}_{i}^{t}\Vert\Vert \textbf{g}_{j}^{t}\Vert}$. Assuming all gradients have bounded norms that $\forall i, \Vert \textbf{g}_{i}^{t}\Vert \leq \epsilon$. The cosine similarity among gradients indicates the coherence: if the gradients have larger cosine similarity, it will have larger descent in the loss and improve the global generalization\footnote{The local gradient coherence is different from gradient diversity \citep{yin2018gradient}. A detailed discussion can be found in \autoref{appendix:analysis_client_coherence} in Appendix}. In this paper, we focus on the local gradient coherence among clients during FL training. We use the cosine stiffness definition \citep{fort2019stiffness} to quantify the local gradient coherence in FL.
\begin{definition} \label{def:local_grad_coherence}
The local gradient coherence of \textit{two clients $i$ and $j$} at round $t$ is defined by the cosine similarity of their local updates sent to the server, as $c_{(i,j)}^{t} = \cos( \textbf{g}_{i}^{t}, \textbf{g}_{j}^{t})$.\\
The overall local gradient coherence of \textit{a cohort of clients} at round $t$ is defined by the weighted cosine similarity of all clients' local updates sent to the server, as $\boldsymbol{c_{cohort}^{t}} = \frac{1}{m}\sum_{i, j, i \neq j}\lambda_{i}\lambda_{j}\cos( \textbf{g}_{i}^{t}, \textbf{g}_{j}^{t})$.
\end{definition}
FL assumes multiple local epochs in each client, and clients usually have heterogeneous data. In this case, the local gradients of clients are usually almost orthogonal, which means that they have low coherence. This phenomenon is observed in \cite{charles2021large}, but it did not dig deeper to examine the training dynamics of FL. In this paper, we calculate the local gradient coherence in each round and find a \textit{critical point} exists in the process (\autoref{fig:hyperplane_noniid} and \autoref{fig:gradient_after_critical}). 

\textbf{Heterogeneity Coherence.} Heterogeneity coherence refers to the distribution consistency between the global data and the sampled one (i.e.\ data distribution of a cohort of sampled clients) in each round. The value of heterogeneity coherence is positively correlated with the IID-ness of clients as well as the client participation ratio; the higher, the better. We define heterogeneity coherence as follows.
\begin{definition} \label{def:hetero_coherence}
Assuming there are $n$ clients and the cohort size is $m$. For a given cohort of clients, the heterogeneity coherence is $\text{\rm sim}(\mathcal{D}_{cohort},\mathcal{D})$, where $\mathcal{D}_{cohort} = \sum_{i \in [m]}\lambda_{i}\mathcal{D}_{i}, \mathcal{D} = \sum_{j = 1}^{n}\lambda_{j}\mathcal{D}_{j}$ and ``{\rm sim}'' is the similarity of two data distributions.
\end{definition}

\subsection{Attentive Learnable Aggregation Weight and Training Dynamics}
\label{subsect:hyperplane1}
Vanilla \textsc{FedAvg} only considers data sizes as clients' aggregation weights $\boldsymbol{\lambda}$. However, clients with different heterogeneity degrees have different importance in client coherence, which can greatly affect the training dynamics. A three-node toy example is shown in \autoref{appdx_fig:toy_example} in Appendix. The optimal $\boldsymbol{\lambda}$ is off the data-sized when clients have the same data size but different heterogeneity degrees. To study client coherence further, we propose attentive learnable aggregation weight (attentive LAW) to learn the optimal aggregation weights (i.e. $\boldsymbol{\lambda}$) on a proxy dataset. By connecting the optimal weights and the client coherence, we can know the roles of different clients in different learning periods. Attentive LAW conducts the model updates in \autoref{eqt_reformulate_fedavg}, where $\{\gamma = 1, \boldsymbol{\lambda}=\boldsymbol{\lambda^{*}}\}$, 

\vspace{-0.3cm}
\begin{small}
\begin{equation} \label{equ:opt_AW_on_plane}
\boldsymbol{\lambda^{*}} = \mathop{\arg\min}\limits_{\boldsymbol{\lambda}}  \mathcal{L}_{proxy} (\sum_{i=1}^{m} \lambda_i\textbf{w}_{i}^{t}), \text{ s.t. } \lambda_{i} \geq 0 , \Vert \boldsymbol{\lambda} \Vert_{1}=1.
\end{equation}
\end{small}
\vspace{-0.2cm}

\begin{figure}[t]
    \centering
     \includegraphics[width=0.7\columnwidth]{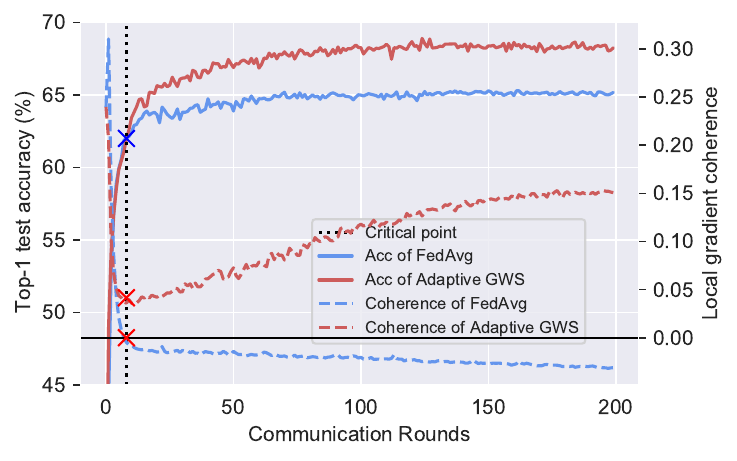}
    \vspace{-1.3em}
    \caption{\small Training dynamics of adaptive GWS in terms of local gradient coherence. $E=3$.}
    \label{fig:gradient_after_critical}
     \vspace{-0.5cm}
\end{figure}

\textbf{1) Generalization will benefit from positive local gradient coherence.} 
\begin{itemize}[leftmargin=*,nosep]
    \item \textbf{Critical point exists in terms of local gradient coherence.} To study the role of client data heterogeneity in local gradient coherence, we experiment on both balanced and imbalanced clients, whose distributions are shown in \autoref{appdx_fig:hyperplane_noniid_distr} of Appendix. 
    The results are demonstrated in \autoref{fig:hyperplane_noniid}, which illustrate that \emph{in the first couple of rounds, the coherence is dominant and positive, thus the test accuracy arises dramatically, and most generalization gains happen in this period}. 
    The critical point is the round that the coherence is near zero. After the critical point, the test accuracy gain is marginal, and the coherence is kept negative but close to zero. 
    \looseness=-1
    \item \textbf{Assigning larger weights to clients with larger coherence before the critical point can improve overall performance.} From the left of \autoref{fig:hyperplane_noniid}, it is clear that before the critical point, the coherence among balanced clients is much higher than that of imbalanced clients. This observation highlights the fact that clients with more balanced data have more coherent gradients
    \footnote{This also reveals why FL performs better in IID settings than NonIID: the clients' gradients in IID settings are more coherent, but the ones in the NonIID usually diverge.}. To capitalize on this, according to \autoref{equ:gradient_coherence}, we can assign larger weights to clients with more balanced data before the critical point to boost generalization. From the right of \autoref{fig:hyperplane_noniid}, attentive LAW proves our hypothesis: it assigns larger weights to balanced clients in the early rounds, particularly in the first two rounds where it nearly assigns all weights to balanced clients.  This may suggest that \emph{the coherence of clients only matters before the critical point where the overall coherence is positive}. 
    To verify this, we adopt early stopping near the critical point when conducting attentive LAW and use data-sized weights after the stopping round.
    Results in the middle of \autoref{fig:hyperplane_noniid} show that \emph{the early-stopped attentive LAW has comparable performance after the critical point.} This insight can guide the design of effective algorithms for learning critically in early training stages.
    \looseness=-1
    \item \textbf{GWS improves local gradient coherence to positive after the critical point.} 
    Interestingly, we observe that \emph{if we adopt adaptive GWS, the local gradient coherence remains positive after the critical point, allowing the model to continue benefiting from the coherent gradients.} As shown in \autoref{fig:gradient_after_critical}, before the critical point, both vanilla \textsc{FedAvg} and adaptive GWS have high gradient coherence, resulting in similar increases in accuracy. After the critical point, the coherence of \textsc{FedAvg} goes down below zero, resulting in marginal performance gains. In contrast, adaptive GWS maintains coherence above zero, allowing for further performance gains beyond \textsc{FedAvg}. 
\end{itemize}

\begin{figure}[!t]
\centering
\includegraphics[width=0.491\columnwidth]{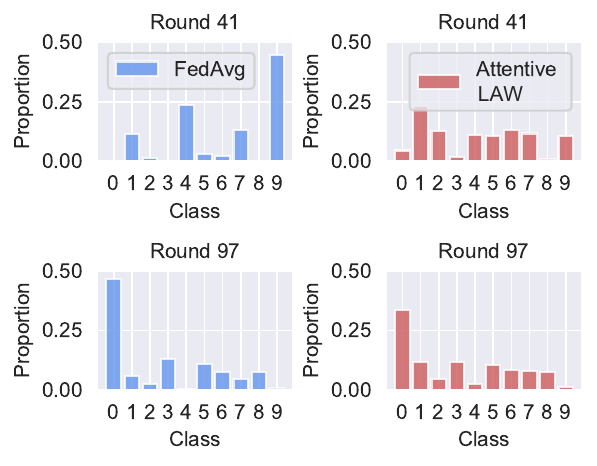}
\includegraphics[width=0.491\columnwidth]{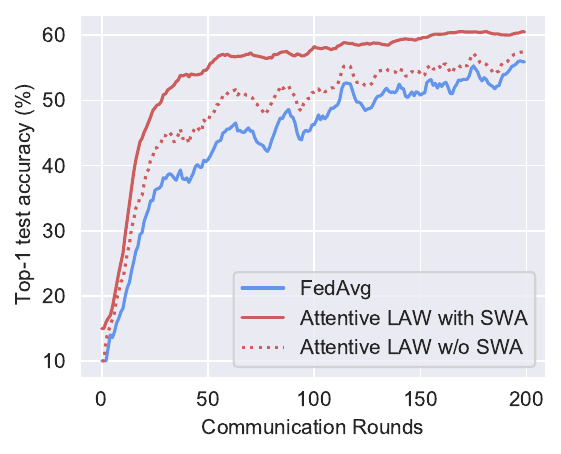}
\vspace{-1.5em}
\caption{\small 
\textbf{Left:} Heterogeneity coherence of class distribution within a cohort. \textbf{Right:} Test accuracy curves. $E=3$.}
\label{fig:hyperplane_partial}
 \vspace{-0.5cm}
\end{figure}

\textbf{2) Improving heterogeneity coherence within a cohort can boost performance.} 
In scenarios with partial client participation in each round, the selected clients have an inconsistent sum objective with the global objective, resulting in low heterogeneity coherence (as defined in \textcolor{darkred}{Definition} \ref{def:hetero_coherence}). 
In theory, the expectation of the sum objective of the sampled clients is consistent with the global objective if the communication rounds (sampling times) are numerous. However, in practical FL, the number of rounds is limited, and from the perspective of learning dynamics, only the first few rounds matter most. Thus, the heterogeneity coherence problem brings a challenge, and sampling methods may not always help, because the availability of clients cannot be guaranteed and stragglers may exist \citep{li2020federated}. 

To address the issue, reweighting the sampled clients in aggregation is quite essential. We find \emph{attentive LAW improves heterogeneity coherence by dynamically adjusting the aggregation weights among clients}. We visualize the weighted class distributions within a cohort in \autoref{fig:hyperplane_partial}, which shows that attentive LAW learns weights to make the class distributions more balanced. 
The test accuracy curves demonstrate a dominant performance gain compared to \textsc{FedAvg}, which showcases the significance of heterogeneity coherence. Additionally, we observe that attentive LAW with SWA\footnote{
Stochastic Weight Averaging (SWA)~\citep{izmailov2018averaging} is an effective technique to make simple averaging of multiple points along the trajectory of optimization with a cyclical learning rate, which leads to better generalization. 
} performs better by seeking a more generalized minimum in the aggregation weight hyperplane. More analysis about client coherence can be found in \autoref{appendix:analysis_client_coherence} of \autoref{appendix:more_results}.

\begin{algorithm}[th] 
  \caption{\textsc{\textbf{FedLAW}}: Federated Learning with Learnable Aggregation Weights}
  \textbf{Input}: {clients $\{1,\dots,n\}$, server-side proxy dataset, communication round $T$, local epoch $E$, server epoch $E_s$, initial global model $\textbf{w}_{g}^{1}$;}\\
  \textbf{Output}: final global model $\textbf{w}_{g}^{T}$;

  \begin{algorithmic}[1]
    \FOR{each round $t=1,\dots, T$}
        \STATE \texttt{\# Client updates}
        \FOR{each client $i, i\in[n]$ \textbf{in parallel}}
        \STATE Set local model $\textbf{w}_i^{t} \leftarrow \textbf{w}_g^{t}$;
        \STATE {Compute $E$ epochs of client local training by \autoref{eqt_client}}:
        \STATE \qquad$\mathbf{w}_{i}^{t}\gets\mathbf{w}_{i}^{t}-\eta_l\nabla \mathcal{L}_{i}\left(\mathbf{w}_{i}^{t}\right)$;
        \ENDFOR
        \STATE \texttt{\# Server updates}
        \STATE The server samples $m$ clients and receive their models $\{\mathbf{w}_{i}^{t}\}_{i=1}^{m}$;
        \STATE The server sets initial $\gamma$ and $\boldsymbol{\lambda}$ as $\{\gamma = 1, \lambda_{i}=\frac{|\mathcal{D}_{i}|}{|\mathcal{D}|}\}$;
        \STATE Compute $E_s$ epochs of aggregation weight learning on the proxy dataset by \autoref{equ:fedawo}:
        \STATE \qquad$\{\gamma,\boldsymbol{\lambda}\}\gets\{\gamma,\boldsymbol{\lambda}\}-\eta_s\nabla \mathcal{L}_{proxy}\left(\{\gamma,\boldsymbol{\lambda}\}\right)$;
        \STATE Obtain the optimal aggregation weights $\{\gamma^*,\boldsymbol{\lambda}^*\}$;
        \STATE Obtain the global model:
        \STATE \qquad $ \textbf{w}_g^{t+1} \leftarrow \gamma^*\cdot(\sum_{i=1}^{m}\lambda^*_i\textbf{w}_{i}^{t})$;
    \ENDFOR
    \STATE Obtain the final global model $\textbf{w}_{g}^{T}$.
  \end{algorithmic}
\label{fedlaw_pseudo-code}
\end{algorithm}

\begin{table*}[t]
    \footnotesize
    \centering
     \vspace{-1.7em}
    \caption{\small \textbf{Top-1 test accuracy (\%) achieved by comparing FL methods and FedLAW on three datasets
    with different model architectures} ($E=3$). 
    \textcolor{blue}{Blue}/\textbf{bold} fonts highlight the best baseline/our approach. 
    }
    \resizebox{\linewidth}{!}{
    \begin{tabular}{c|cc|cc|cc|cc|cc|cc}
    \toprule
    Dataset&\multicolumn{4}{c}{FashionMNIST}&\multicolumn{4}{c}{CIFAR-10}&\multicolumn{4}{c}{CIFAR-100}\\
    \cmidrule(lr){1-5}
    \cmidrule(lr){6-9}
    \cmidrule(lr){10-13}
    NonIID ($\alpha$) &\multicolumn{2}{c}{100}&\multicolumn{2}{c}{0.1}&\multicolumn{2}{c}{100}&\multicolumn{2}{c}{0.1}&\multicolumn{2}{c}{100}&\multicolumn{2}{c}{0.1}\\
    \midrule
    Model &MLP&LeNet&MLP&LeNet&CNN&ResNet&CNN&ResNet&CNN&ResNet&CNN&ResNet\\
    \midrule
    \textsc{FedAvg}&\textcolor{blue}{89.29}	&90.54	&85.11	&88.08
    &65.78	&74.57	&60.13	&46.04	&25.74	&27.49	&\textcolor{blue}{27.74}	&24.92\\
    \midrule
    \textsc{FedProx}&87.68	&89.77	&84.33	&87.01
    &67.66	&68.51	&\textcolor{blue}{60.48}	&48.84	&9.49	&27.15	&12.52	&23.73\\
    \textsc{FedDyn}&88.47	&89.92	&77.68	&72.68
    &66.1	&76.62	&41.53	&35.77	&24.44	&\textcolor{blue}{32.18}	&22.67	&\textcolor{blue}{29.00}\\
    \midrule
    \textsc{FedDF}&86.16	&89.09	&78.48	&85.90
    &69.60	&77.36	&57.38	&\textcolor{blue}{54.09}	&\textcolor{blue}{28.52}	&27.42	&24.52	&23.10\\
    \textsc{FedBE}&86.22	&89.14	&79.12	&85.96
    &\textcolor{blue}{69.88} &\textcolor{blue}{77.94}	&59.84	&52.86	&28.38	&27.73	&25.41	&23.74\\
    \textsc{Server-FT}&89.09	&\textcolor{blue}{90.56}	&\textcolor{blue}{85.71}	&\textcolor{blue}{88.10}
    &66.83 &74.73	&60.43	&47.59	&25.37	&26.14	&24.33	&23.03\\
    \midrule
    \textbf{\textsc{FedLAW}}&\textbf{88.51}	&\textbf{90.66}	&\textbf{86.30}	&\textbf{88.26}
    &\textbf{70.17}	&\textbf{80.46}	&\textbf{62.46}	&\textbf{52.83}	&\textbf{32.51}	&\textbf{33.17}	&\textbf{32.30}	&\textbf{24.84}\\
    \textbf{\textsc{FedLAW (SWA)}}&\textbf{88.27} &\textbf{90.51}	&\textbf{86.89}	&\textbf{88.18}	&\textbf{69.9}	&\textbf{79.55}	&\textbf{62.12}	&\textbf{57.08}	&\textbf{32.39}	&\textbf{33.17}	&\textbf{32.27}	&\textbf{25.31}\\
    \bottomrule
    \end{tabular}
    }
    \label{table:first_table}
    \vspace{-0.2em}
\end{table*}

\section{FedLAW} \label{sect:fedawo}
\subsection{Method}
Based on the above understandings, we propose \textbf{Fed}erated Learning with \textbf{L}earnable \textbf{A}ggregation \textbf{W}eights algorithm (\textsc{\textbf{FedLAW}}) which combines the adaptive GWS and attentive LAW to optimize $\gamma$ and $\boldsymbol{\lambda}$ simultaneously, defined as
\begin{align}
 \label{equ:fedawo}
 \boldsymbol{\gamma^{*},\lambda^{*}} &= \mathop{\arg\min}\limits_{{\gamma,\boldsymbol{\lambda}}}  \mathcal{L}_{proxy} \gamma\cdot(\sum_{i=1}^{m}\lambda_i\textbf{w}_{i}^{t}),\\ &\text{ s.t. } \gamma > 0,\lambda_{i} \geq 0 , \Vert \boldsymbol{\lambda} \Vert_{1} = 1.
\end{align}
The pseudo-code of \textsc{FedLAW} is shown in \textcolor{darkred}{Algorithm} \ref{fedlaw_pseudo-code}.

\textbf{With SWA (optional).} We adopt an alternative two-stage strategy for SWA variant (implementing it in a reversed order also works), where we first fix $\boldsymbol{\lambda}$ and optimize $\gamma$, then we use the learned $\gamma$ and fix it to optimize $\boldsymbol{\lambda}$ with SWA. 

In our experiments, we denote \textsc{FedLAW} with or without SWA as ``\textsc{FedLAW (SWA)}'' or ``\textsc{FedLAW}''.

\begin{table}[]
\centering
\vspace{-0.4cm}
\caption{Performance comparison under different numbers of clients. CIFAR-10, ResNet20, $E=3$.}
\resizebox{0.8\linewidth}{!}{
\begin{tabular}{@{}c|cccc@{}}
\toprule
               Setting    & \multicolumn{2}{c}{IID ($\alpha=100$)}                    & \multicolumn{2}{c}{NonIID ($\alpha=1$)}                   \\\midrule
Number of clients $n$  & 50 & 100 & 50 & 100             \\ \midrule
\textsc{FedAvg}             & 68.04            & 62.41          & 66.87          & 64.13          \\
\textsc{FedDF}              & 48.24            & 38.66          & 38.70          & 22.51  \\
\textsc{Server-FT}              & 67.77            & 62.30          & 66.73          & 64.63  \\
\midrule
\textsc{FedLAW}             & \textbf{78.88}   & \textbf{74.09} & \textbf{75.59} & \textbf{71.34}  \\ \bottomrule
\end{tabular}
}
\label{tab:client_num}
\vspace{-0.2cm}
\end{table}

\begin{table}[t]
    \centering
    \vspace{-0.2cm} 
	\tiny
	\caption{\small The performance of compared methods with different model architectures ($\alpha=1,~E=1$).} 
    \resizebox{0.9\linewidth}{!}{
	\begin{tabular}{l|ccc}
	\toprule
	Model & \textsc{FedAvg} &  \textsc{FedLAW} & \textsc{FedLAW(SWA)}\\
    \midrule
    ResNet20 & 74.11 & 78.72 & 78.64 \\ 
    ResNet56 & 74.22  & 78.93 & 79.08 \\ 
    ResNet110 & 74.50  & 78.11 & 79.19\\ 
    \midrule
    WRN56\_4& 78.67  & 79.61  & 80.70\\
    \midrule
    DenseNet121& 85.13  & 86.50 & 87.06\\
    \bottomrule
	\end{tabular}}
	\label{table:model_architetures}
 \vspace{-0.2cm}
\end{table}

\begin{figure*}[t]
    \centering
    \centering
    \includegraphics[width=0.85\columnwidth]{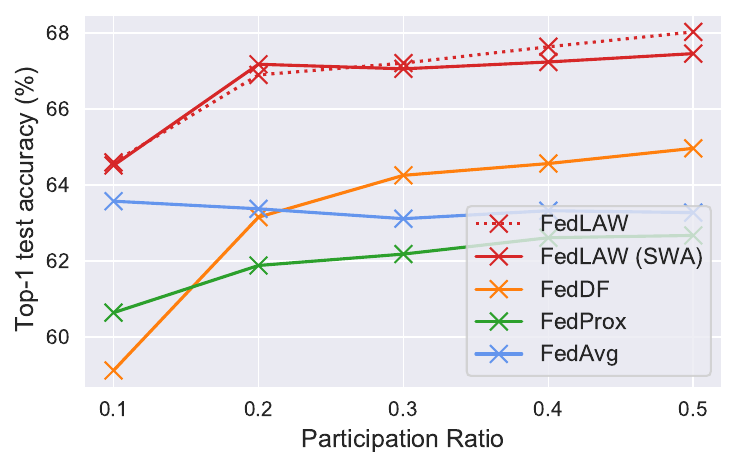}
    \includegraphics[width=0.85\columnwidth]{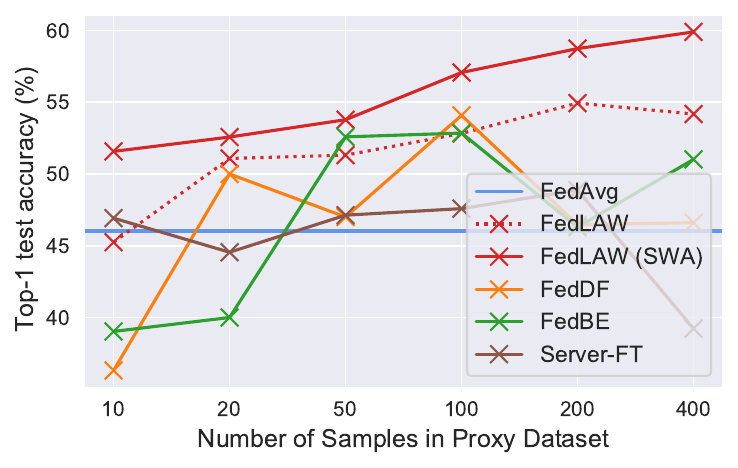}
    \vspace{-1.5em}
    \caption{\small \textbf{Left:} The performance with different participation ratios ($\alpha=1,~E=3$). \textbf{Right:} The performance with different sizes of the proxy dataset ($\alpha=0.1,~E=3$).
    }
    \label{fig:fedawo_partial}
     \vspace{-0.4cm}
\end{figure*}

\subsection{Experiments}
\textbf{Baselines and Settings.} We conduct experiments to verify the effectiveness of \textsc{FedLAW}. We mainly compare \textsc{FedLAW} with other server-side methods, i.e.\ \textsc{FedDF} \citep{lin2020ensemble} and \textsc{FedBE} \citep{DBLP:conf/iclr/ChenC21}, that also require a proxy dataset for additional computation. These two methods conduct ensemble distillation on the proxy data to transfer knowledge from clients' models to the global model. We add \textsc{Server-FT} as a baseline for simply finetuning global models on the proxy dataset. Besides, we implement client-side algorithms \textsc{FedProx} \citep{li2020federated} and \textsc{FedDyn} \citep{acar2020federated} for comparison. If not mentioned otherwise, the number of clients is 20. More implementation details can be found in \autoref{appendix:detail_fedawo} and \autoref{appendix:detail_implementation}.

\textbf{Experimental results. \textit{Different datasets:}} As in \autoref{table:first_table}, \textsc{FedLAW} outperforms baselines on different datasets and models in both IID and NonIID settings. Compared with \textsc{FedDF}, \textsc{FedBE} and \textsc{Server-FT}, \textsc{FedLAW} can better utilize the proxy dataset. 
\textbf{\textit{Different numbers of clients:}} We implement experiments by scaling up the number of clients in \autoref{tab:client_num}, and it is shown that \textsc{FedLAW} also surpasses the baselines by large margins. 
\textbf{\textit{Different model architectures:}} We test \textsc{FedLAW} across wider and deeper ResNet and other architecture, such as DenseNet \citep{huang2017densely}, in the \autoref{table:model_architetures}. It shows that \textsc{FedLAW} is effective across different architectures, and it performs well even when the network goes deeper or wider. 
\textbf{\textit{Different participation ratios:}} From the left of \autoref{fig:fedawo_partial}, \textsc{FedLAW} performs well under partial participation. 
\textbf{\textit{Different sizes and distributions of proxy dataset:}} From the right of \autoref{fig:fedawo_partial}, the server-side baselines are sensitive to the size of the proxy dataset that too small or too large proxy set will cause overfitting. However, \textsc{FedLAW} is also effective under an extremely tiny proxy set and benefits more from a larger proxy set due to accurate aggregation weight optimization. We report the results of different distributions of the proxy dataset in \autoref{tab:distribution_shift_2} and \autoref{tab:distribution_shift_1}, which show that FedLAW still works when there exists a distribution shift between the proxy dataset and the global data distribution of clients. 
\textbf{\textit{Robustness against corrupted clients:}} Another advantage of \textsc{FedLAW} is that it can filter out corrupted clients by assigning them lower weights. We generate corrupted clients by swapping two labels in their local training data. As in \autoref{table:robustness}, \textsc{FedLAW} is robust against corrupted clients, and it is as robust as the ensemble distillation methods, such as \textsc{FedDF}, using the same proxy dataset.

\textbf{More results.} We present more results in the appendix. Specifically, the learning curves of test accuracy (\textcolor{darkred}{Figures} \ref{appdx_fig:curve_fmnist}-\ref{appdx_fig:curve_CIFAR-100}) and the server training process of \textsc{FedLAW} (\autoref{appdx_fig:server_training_vis}).

\begin{table}[]
\centering
\vspace{-0.3cm}
\caption{The performance on the distribution shift setting where the clients' data are overall balanced and the proxy data are long-tailed ($\rho=10$).}
\resizebox{1.0\linewidth}{!}{
\begin{tabular}{@{}c|cccccc@{}}
\toprule
               Setting    & \multicolumn{3}{c}{IID ($\alpha=100$)}                    & \multicolumn{3}{c}{NonIID ($\alpha=1$)}                   \\\midrule
Type of proxy data & Balanced         & \multicolumn{2}{c}{Long-tailed} & Balanced         & \multicolumn{2}{c}{Long-tailed} \\ 
Balanced Sampling  & - & w/o            & w              & - & w/o            & w              \\ \midrule
\textsc{FedAvg}             & 75.24            & 75.24          & 75.24          & 73.46            & 73.46          & \textbf{73.46} \\
\textsc{FedDF}              & 76.20            & 74.04          & 73.31          & 74.39            & 73.99          & 73.14          \\\midrule
\textsc{FedLAW}             & \textbf{79.40}   & \textbf{77.14} & \textbf{78.56} & \textbf{76.70}   & \textbf{76.78} & 70.42          \\ \bottomrule
\end{tabular}
}
\label{tab:distribution_shift_2}
\vspace{-0.2cm}
\end{table}

\begin{table}[!t] 
	\centering
 \vspace{-0.15cm}
        \begin{minipage}{0.45\linewidth}
        \centering
        \caption{\small The performance on the distribution shift setting where the clients' data are long-tailed ($\rho=5$) and the proxy data are balanced.}
        \resizebox{\linewidth}{!}{
        \begin{tabular}{@{}c|cc@{}}
        \toprule
            Method & IID ($\alpha=100$) & NonIID ($\alpha=1$) \\ \midrule
            \textsc{FedAvg} & 61.12       & 59.82        \\
            \textsc{FedDF}  & 39.03       & 40.68        \\ \midrule
            \textsc{FedLAW} & \textbf{67.61}       & \textbf{66.48}        \\ \bottomrule
        \end{tabular}
        }
        \label{tab:distribution_shift_1}
        \end{minipage}
    \hfill
        \begin{minipage}{0.5\linewidth}
            \caption{\small The performance on different percentages of corrupted clients (IID, $E=3$).}
            \resizebox{\linewidth}{!}{
            \begin{tabular}{l|ccc}
            	\toprule
            	Corrupt percent. & 25\% &  50\% & 75\%\\
                \midrule
                \textsc{FedAvg} & 63.40 & 61.14 & 58.21 \\
                \midrule
                \textsc{FedDF} & 68.73  & 66.94 & 66.07 \\
                \textsc{Server-FT} & 63.61  & 61.24 & 58.36\\
                \midrule
                \textsc{FedLAW} & 67.87  & 66.67 & 63.91\\
                \textsc{FedLAW(SWA)} & 68.04  & 66.95 & 65.51\\
                \bottomrule
        	\end{tabular}}
        	\label{table:robustness}
        \end{minipage}
\vspace{-0.08cm}
\end{table}

\section{Conclusion} \vspace{0.15cm}
In this paper, we revisit and rethink the weighted aggregation in federated learning with neural networks and gain new insights into the training dynamics. First, we break the convention that the $l_1$ norm of aggregation weights should be normalized as 1 and identify the \textit{global weight shrinking} phenomenon and its dynamics when the norm is smaller than 1. 
Second, we discover two aspects of \textit{client coherence}, local gradient coherence and heterogeneity coherence, and study the dynamics during training. 
Based on the findings, we devise a simple but effective method \textsc{FedLAW}. 
Extensive experiments verify that our method can improve the generalization of the global model by a large margin on different datasets and models. 


\section*{Acknowledgments}
This work was supported by the National Key Research and Development Project of China (Grant No.2021ZD0110505), the National Natural Science Foundation of China (Grant No.U19B2042), the Zhejiang Provincial Key Research and Development Project (Grant No.2022C01044), the University Synergy Innovation Program of Anhui Province (Grant No.GXXT-2021-004), the Academy Of Social Governance Zhejiang University, and the Fundamental Research Funds for the Central Universities (Grant No.226-2022-00064). This work was also supported in part by the Research Center for Industries of the Future (RCIF) at Westlake University, and Westlake Education Foundation.


\newpage
\bibliography{icml2023}

\begin{thebibliography}{58}
\providecommand{\natexlab}[1]{#1}
\providecommand{\url}[1]{\texttt{#1}}
\expandafter\ifx\csname urlstyle\endcsname\relax
  \providecommand{\doi}[1]{doi: #1}\else
  \providecommand{\doi}{doi: \begingroup \urlstyle{rm}\Url}\fi

\bibitem[Acar et~al.(2020)Acar, Zhao, Matas, Mattina, Whatmough, and
  Saligrama]{acar2020federated}
Acar, D. A.~E., Zhao, Y., Matas, R., Mattina, M., Whatmough, P., and Saligrama,
  V.
\newblock Federated learning based on dynamic regularization.
\newblock In \emph{International Conference on Learning Representations}, 2020.

\bibitem[Allen-Zhu et~al.(2019)Allen-Zhu, Li, and Song]{allen2019convergence}
Allen-Zhu, Z., Li, Y., and Song, Z.
\newblock A convergence theory for deep learning via over-parameterization.
\newblock In \emph{International Conference on Machine Learning}, pp.\
  242--252. PMLR, 2019.

\bibitem[Caldarola et~al.(2022)Caldarola, Caputo, and
  Ciccone]{caldarola2022improving}
Caldarola, D., Caputo, B., and Ciccone, M.
\newblock Improving generalization in federated learning by seeking flat
  minima.
\newblock In \emph{Computer Vision--ECCV 2022: 17th European Conference, Tel
  Aviv, Israel, October 23--27, 2022, Proceedings, Part XXIII}, pp.\  654--672.
  Springer, 2022.

\bibitem[Charles et~al.(2021)Charles, Garrett, Huo, Shmulyian, and
  Smith]{charles2021large}
Charles, Z., Garrett, Z., Huo, Z., Shmulyian, S., and Smith, V.
\newblock On large-cohort training for federated learning.
\newblock \emph{Advances in neural information processing systems},
  34:\penalty0 20461--20475, 2021.

\bibitem[Chatterjee(2019)]{chatterjee2019coherent}
Chatterjee, S.
\newblock Coherent gradients: An approach to understanding generalization in
  gradient descent-based optimization.
\newblock In \emph{International Conference on Learning Representations}, 2019.

\bibitem[Chatterjee \& Zielinski(2020)Chatterjee and
  Zielinski]{chatterjee2020making}
Chatterjee, S. and Zielinski, P.
\newblock Making coherence out of nothing at all: measuring the evolution of
  gradient alignment.
\newblock \emph{arXiv preprint arXiv:2008.01217}, 2020.

\bibitem[Chen \& Chao(2021)Chen and Chao]{DBLP:conf/iclr/ChenC21}
Chen, H. and Chao, W.
\newblock Fedbe: Making bayesian model ensemble applicable to federated
  learning.
\newblock In \emph{9th International Conference on Learning Representations,
  {ICLR} 2021, Virtual Event, Austria, May 3-7, 2021}. OpenReview.net, 2021.
\newblock URL \url{https://openreview.net/forum?id=dgtpE6gKjHn}.

\bibitem[Deng et~al.(2020)Deng, Kamani, and Mahdavi]{deng2020distributionally}
Deng, Y., Kamani, M.~M., and Mahdavi, M.
\newblock Distributionally robust federated averaging.
\newblock \emph{Advances in neural information processing systems},
  33:\penalty0 15111--15122, 2020.

\bibitem[Dinh et~al.(2017)Dinh, Pascanu, Bengio, and Bengio]{dinh2017sharp}
Dinh, L., Pascanu, R., Bengio, S., and Bengio, Y.
\newblock Sharp minima can generalize for deep nets.
\newblock In \emph{International Conference on Machine Learning}, pp.\
  1019--1028. PMLR, 2017.

\bibitem[Draxler et~al.(2018)Draxler, Veschgini, Salmhofer, and
  Hamprecht]{draxler2018essentially}
Draxler, F., Veschgini, K., Salmhofer, M., and Hamprecht, F.
\newblock Essentially no barriers in neural network energy landscape.
\newblock In \emph{International conference on machine learning}, pp.\
  1309--1318. PMLR, 2018.

\bibitem[Du et~al.(2021)Du, Yan, Feng, Zhou, Zhen, Goh, and
  Tan]{du2021efficient}
Du, J., Yan, H., Feng, J., Zhou, J.~T., Zhen, L., Goh, R. S.~M., and Tan, V.~Y.
\newblock Efficient sharpness-aware minimization for improved training of
  neural networks.
\newblock \emph{arXiv preprint arXiv:2110.03141}, 2021.

\bibitem[Entezari et~al.(2022)Entezari, Sedghi, Saukh, and
  Neyshabur]{entezari2021role}
Entezari, R., Sedghi, H., Saukh, O., and Neyshabur, B.
\newblock The role of permutation invariance in linear mode connectivity of
  neural networks.
\newblock In \emph{International Conference on Learning Representations}, 2022.

\bibitem[Foret et~al.(2020)Foret, Kleiner, Mobahi, and
  Neyshabur]{foret2020sharpness}
Foret, P., Kleiner, A., Mobahi, H., and Neyshabur, B.
\newblock Sharpness-aware minimization for efficiently improving
  generalization.
\newblock \emph{arXiv preprint arXiv:2010.01412}, 2020.

\bibitem[Fort \& Jastrzebski(2019)Fort and Jastrzebski]{fort2019large}
Fort, S. and Jastrzebski, S.
\newblock Large scale structure of neural network loss landscapes.
\newblock \emph{Advances in Neural Information Processing Systems}, 32, 2019.

\bibitem[Fort et~al.(2019)Fort, Nowak, Jastrzebski, and
  Narayanan]{fort2019stiffness}
Fort, S., Nowak, P.~K., Jastrzebski, S., and Narayanan, S.
\newblock Stiffness: A new perspective on generalization in neural networks.
\newblock \emph{arXiv preprint arXiv:1901.09491}, 2019.

\bibitem[Franceschi et~al.(2017)Franceschi, Donini, Frasconi, and
  Pontil]{franceschi2017forward}
Franceschi, L., Donini, M., Frasconi, P., and Pontil, M.
\newblock Forward and reverse gradient-based hyperparameter optimization.
\newblock In \emph{International Conference on Machine Learning}, pp.\
  1165--1173. PMLR, 2017.

\bibitem[Garipov et~al.(2018)Garipov, Izmailov, Podoprikhin, Vetrov, and
  Wilson]{garipov2018loss}
Garipov, T., Izmailov, P., Podoprikhin, D., Vetrov, D.~P., and Wilson, A.~G.
\newblock Loss surfaces, mode connectivity, and fast ensembling of dnns.
\newblock \emph{Advances in neural information processing systems}, 31, 2018.

\bibitem[Guo et~al.(2022)Guo, Yang, Hatamizadeh, Xu, Xu, Li, Zhao, Xu, Harmon,
  Turkbey, et~al.]{guo2022auto}
Guo, P., Yang, D., Hatamizadeh, A., Xu, A., Xu, Z., Li, W., Zhao, C., Xu, D.,
  Harmon, S., Turkbey, E., et~al.
\newblock Auto-fedrl: Federated hyperparameter optimization for
  multi-institutional medical image segmentation.
\newblock \emph{arXiv preprint arXiv:2203.06338}, 2022.

\bibitem[Huang et~al.(2017)Huang, Liu, Van Der~Maaten, and
  Weinberger]{huang2017densely}
Huang, G., Liu, Z., Van Der~Maaten, L., and Weinberger, K.~Q.
\newblock Densely connected convolutional networks.
\newblock In \emph{Proceedings of the IEEE conference on computer vision and
  pattern recognition}, pp.\  4700--4708, 2017.

\bibitem[Huang et~al.(2021)Huang, Chu, Zhou, Wang, Liu, Pei, and
  Zhang]{huang2021personalized}
Huang, Y., Chu, L., Zhou, Z., Wang, L., Liu, J., Pei, J., and Zhang, Y.
\newblock Personalized cross-silo federated learning on non-iid data.
\newblock In \emph{Proceedings of the AAAI Conference on Artificial
  Intelligence}, volume~35, pp.\  7865--7873, 2021.

\bibitem[Izmailov et~al.(2018)Izmailov, Wilson, Podoprikhin, Vetrov, and
  Garipov]{izmailov2018averaging}
Izmailov, P., Wilson, A., Podoprikhin, D., Vetrov, D., and Garipov, T.
\newblock Averaging weights leads to wider optima and better generalization.
\newblock In \emph{34th Conference on Uncertainty in Artificial Intelligence
  2018, UAI 2018}, pp.\  876--885, 2018.

\bibitem[Jastrz{\k{e}}bski et~al.(2018)Jastrz{\k{e}}bski, Kenton, Ballas,
  Fischer, Bengio, and Storkey]{jastrzkebski2018relation}
Jastrz{\k{e}}bski, S., Kenton, Z., Ballas, N., Fischer, A., Bengio, Y., and
  Storkey, A.
\newblock On the relation between the sharpest directions of dnn loss and the
  sgd step length.
\newblock \emph{arXiv preprint arXiv:1807.05031}, 2018.

\bibitem[Jastrzebski et~al.(2019)Jastrzebski, Szymczak, Fort, Arpit, Tabor,
  Cho, and Geras]{jastrzebski2019break}
Jastrzebski, S., Szymczak, M., Fort, S., Arpit, D., Tabor, J., Cho, K., and
  Geras, K.
\newblock The break-even point on optimization trajectories of deep neural
  networks.
\newblock In \emph{International Conference on Learning Representations}, 2019.

\bibitem[Jastrzebski et~al.(2020)Jastrzebski, Szymczak, Fort, Arpit, Tabor,
  Cho, and Geras]{jastrzebski2020break}
Jastrzebski, S., Szymczak, M., Fort, S., Arpit, D., Tabor, J., Cho, K., and
  Geras, K.
\newblock The break-even point on optimization trajectories of deep neural
  networks.
\newblock \emph{arXiv preprint arXiv:2002.09572}, 2020.

\bibitem[Jomaa et~al.(2019)Jomaa, Grabocka, and Schmidt-Thieme]{jomaa2019hyp}
Jomaa, H.~S., Grabocka, J., and Schmidt-Thieme, L.
\newblock Hyp-rl: Hyperparameter optimization by reinforcement learning.
\newblock \emph{arXiv preprint arXiv:1906.11527}, 2019.

\bibitem[Kantorovich(2006)]{kantorovich2006translocation}
Kantorovich, L.~V.
\newblock On the translocation of masses.
\newblock \emph{Journal of mathematical sciences}, 133\penalty0 (4):\penalty0
  1381--1382, 2006.

\bibitem[Karimireddy et~al.(2020)Karimireddy, Kale, Mohri, Reddi, Stich, and
  Suresh]{karimireddy2020scaffold}
Karimireddy, S.~P., Kale, S., Mohri, M., Reddi, S., Stich, S., and Suresh,
  A.~T.
\newblock Scaffold: Stochastic controlled averaging for federated learning.
\newblock In \emph{International Conference on Machine Learning}, pp.\
  5132--5143. PMLR, 2020.

\bibitem[Keskar et~al.(2017)Keskar, Nocedal, Tang, Mudigere, and
  Smelyanskiy]{keskar2017large}
Keskar, N.~S., Nocedal, J., Tang, P. T.~P., Mudigere, D., and Smelyanskiy, M.
\newblock On large-batch training for deep learning: Generalization gap and
  sharp minima.
\newblock In \emph{5th International Conference on Learning Representations,
  ICLR 2017}, 2017.

\bibitem[Kwon et~al.(2021)Kwon, Kim, Park, and Choi]{kwon2021asam}
Kwon, J., Kim, J., Park, H., and Choi, I.~K.
\newblock Asam: Adaptive sharpness-aware minimization for scale-invariant
  learning of deep neural networks.
\newblock In \emph{International Conference on Machine Learning}, pp.\
  5905--5914. PMLR, 2021.

\bibitem[Lewkowycz \& Gur-Ari(2020)Lewkowycz and
  Gur-Ari]{lewkowycz2020training}
Lewkowycz, A. and Gur-Ari, G.
\newblock On the training dynamics of deep networks with $ l\_2 $
  regularization.
\newblock \emph{Advances in Neural Information Processing Systems},
  33:\penalty0 4790--4799, 2020.

\bibitem[Li et~al.(2018)Li, Xu, Taylor, Studer, and
  Goldstein]{li2018visualizing}
Li, H., Xu, Z., Taylor, G., Studer, C., and Goldstein, T.
\newblock Visualizing the loss landscape of neural nets.
\newblock \emph{Advances in neural information processing systems}, 31, 2018.

\bibitem[Li et~al.(2022{\natexlab{a}})Li, Zhou, Tian, and Tao]{li2022learning}
Li, S., Zhou, T., Tian, X., and Tao, D.
\newblock Learning to collaborate in decentralized learning of personalized
  models.
\newblock In \emph{Proceedings of the IEEE/CVF Conference on Computer Vision
  and Pattern Recognition}, pp.\  9766--9775, 2022{\natexlab{a}}.

\bibitem[Li et~al.(2020{\natexlab{a}})Li, Sahu, Talwalkar, and
  Smith]{DBLP:journals/spm/LiSTS20}
Li, T., Sahu, A.~K., Talwalkar, A., and Smith, V.
\newblock Federated learning: Challenges, methods, and future directions.
\newblock \emph{{IEEE} Signal Process. Mag.}, 37\penalty0 (3):\penalty0 50--60,
  2020{\natexlab{a}}.
\newblock \doi{10.1109/MSP.2020.2975749}.
\newblock URL \url{https://doi.org/10.1109/MSP.2020.2975749}.

\bibitem[Li et~al.(2020{\natexlab{b}})Li, Sahu, Zaheer, Sanjabi, Talwalkar, and
  Smith]{li2020federated}
Li, T., Sahu, A.~K., Zaheer, M., Sanjabi, M., Talwalkar, A., and Smith, V.
\newblock Federated optimization in heterogeneous networks.
\newblock \emph{Proceedings of Machine Learning and Systems}, 2:\penalty0
  429--450, 2020{\natexlab{b}}.

\bibitem[Li et~al.(2020{\natexlab{c}})Li, Chen, and Yang]{li2020understanding}
Li, X., Chen, S., and Yang, J.
\newblock Understanding the disharmony between weight normalization family and
  weight decay.
\newblock In \emph{Proceedings of the AAAI Conference on Artificial
  Intelligence}, volume~34, pp.\  4715--4722, 2020{\natexlab{c}}.

\bibitem[Li et~al.(2022{\natexlab{b}})Li, Lu, Luo, Zhu, Shao, Li, Zhang, Wang,
  and Wu]{li2022towards}
Li, Z., Lu, J., Luo, S., Zhu, D., Shao, Y., Li, Y., Zhang, Z., Wang, Y., and
  Wu, C.
\newblock Towards effective clustered federated learning: A peer-to-peer
  framework with adaptive neighbor matching.
\newblock \emph{IEEE Transactions on Big Data}, 2022{\natexlab{b}}.

\bibitem[Li et~al.(2022{\natexlab{c}})Li, Lu, Luo, Zhu, Shao, Li, Zhang, and
  Wu]{li2022mining}
Li, Z., Lu, J., Luo, S., Zhu, D., Shao, Y., Li, Y., Zhang, Z., and Wu, C.
\newblock Mining latent relationships among clients: Peer-to-peer federated
  learning with adaptive neighbor matching.
\newblock \emph{arXiv preprint arXiv:2203.12285}, 2022{\natexlab{c}}.

\bibitem[Lin et~al.(2020)Lin, Kong, Stich, and Jaggi]{lin2020ensemble}
Lin, T., Kong, L., Stich, S.~U., and Jaggi, M.
\newblock Ensemble distillation for robust model fusion in federated learning.
\newblock \emph{Advances in Neural Information Processing Systems},
  33:\penalty0 2351--2363, 2020.

\bibitem[Loshchilov \& Hutter(2018)Loshchilov and
  Hutter]{loshchilov2018decoupled}
Loshchilov, I. and Hutter, F.
\newblock Decoupled weight decay regularization.
\newblock In \emph{International Conference on Learning Representations}, 2018.

\bibitem[Lyu et~al.(2022)Lyu, Li, and Arora]{lyu2022understanding}
Lyu, K., Li, Z., and Arora, S.
\newblock Understanding the generalization benefit of normalization layers:
  Sharpness reduction.
\newblock \emph{arXiv preprint arXiv:2206.07085}, 2022.

\bibitem[Maclaurin et~al.(2015)Maclaurin, Duvenaud, and
  Adams]{maclaurin2015gradient}
Maclaurin, D., Duvenaud, D., and Adams, R.
\newblock Gradient-based hyperparameter optimization through reversible
  learning.
\newblock In \emph{International conference on machine learning}, pp.\
  2113--2122. PMLR, 2015.

\bibitem[McMahan et~al.(2017)McMahan, Moore, Ramage, Hampson, and
  y~Arcas]{mcmahan2017communication}
McMahan, B., Moore, E., Ramage, D., Hampson, S., and y~Arcas, B.~A.
\newblock Communication-efficient learning of deep networks from decentralized
  data.
\newblock In \emph{Artificial intelligence and statistics}, pp.\  1273--1282.
  PMLR, 2017.

\bibitem[Mostafa(2019)]{mostafa2019robust}
Mostafa, H.
\newblock Robust federated learning through representation matching and
  adaptive hyper-parameters.
\newblock \emph{arXiv preprint arXiv:1912.13075}, 2019.

\bibitem[Singh \& Jaggi(2020)Singh and Jaggi]{singh2020model}
Singh, S.~P. and Jaggi, M.
\newblock Model fusion via optimal transport.
\newblock \emph{Advances in Neural Information Processing Systems},
  33:\penalty0 22045--22055, 2020.

\bibitem[Vlaar \& Frankle(2021)Vlaar and Frankle]{vlaar2021can}
Vlaar, T. and Frankle, J.
\newblock What can linear interpolation of neural network loss landscapes tell
  us?
\newblock \emph{arXiv preprint arXiv:2106.16004}, 2021.

\bibitem[Wan et~al.(2021)Wan, Zhu, Zhang, and Sun]{wan2021spherical}
Wan, R., Zhu, Z., Zhang, X., and Sun, J.
\newblock Spherical motion dynamics: Learning dynamics of normalized neural
  network using sgd and weight decay.
\newblock \emph{Advances in Neural Information Processing Systems},
  34:\penalty0 6380--6391, 2021.

\bibitem[Wang et~al.(2020{\natexlab{a}})Wang, Yurochkin, Sun, Papailiopoulos,
  and Khazaeni]{wang2020federated}
Wang, H., Yurochkin, M., Sun, Y., Papailiopoulos, D., and Khazaeni, Y.
\newblock Federated learning with matched averaging.
\newblock \emph{arXiv preprint arXiv:2002.06440}, 2020{\natexlab{a}}.

\bibitem[Wang et~al.(2020{\natexlab{b}})Wang, Liu, Liang, Joshi, and
  Poor]{wang2020tackling}
Wang, J., Liu, Q., Liang, H., Joshi, G., and Poor, H.~V.
\newblock Tackling the objective inconsistency problem in heterogeneous
  federated optimization.
\newblock \emph{Advances in neural information processing systems},
  33:\penalty0 7611--7623, 2020{\natexlab{b}}.

\bibitem[Wang et~al.(2021)Wang, Charles, Xu, Joshi, McMahan, Al-Shedivat,
  Andrew, Avestimehr, Daly, Data, et~al.]{wang2021field}
Wang, J., Charles, Z., Xu, Z., Joshi, G., McMahan, H.~B., Al-Shedivat, M.,
  Andrew, G., Avestimehr, S., Daly, K., Data, D., et~al.
\newblock A field guide to federated optimization.
\newblock \emph{arXiv preprint arXiv:2107.06917}, 2021.

\bibitem[Wu et~al.(2022)Wu, Liang, Han, Bian, Zhao, and Huang]{wu2022drflm}
Wu, B., Liang, Z., Han, Y., Bian, Y., Zhao, P., and Huang, J.
\newblock Drflm: Distributionally robust federated learning with inter-client
  noise via local mixup.
\newblock \emph{arXiv preprint arXiv:2204.07742}, 2022.

\bibitem[Xia et~al.(2021)Xia, Yang, Li, Myronenko, Xu, Obinata, Mori, An,
  Harmon, Turkbey, et~al.]{xia2021auto}
Xia, Y., Yang, D., Li, W., Myronenko, A., Xu, D., Obinata, H., Mori, H., An,
  P., Harmon, S., Turkbey, E., et~al.
\newblock Auto-fedavg: learnable federated averaging for multi-institutional
  medical image segmentation.
\newblock \emph{arXiv preprint arXiv:2104.10195}, 2021.

\bibitem[Xie et~al.(2020)Xie, Sato, and Sugiyama]{xie2020understanding}
Xie, Z., Sato, I., and Sugiyama, M.
\newblock Understanding and scheduling weight decay.
\newblock \emph{arXiv preprint arXiv:2011.11152}, 2020.

\bibitem[Yan et~al.(2021)Yan, Wang, and Li]{yan2021critical}
Yan, G., Wang, H., and Li, J.
\newblock Critical learning periods in federated learning.
\newblock \emph{arXiv preprint arXiv:2109.05613}, 2021.

\bibitem[Yao et~al.(2020)Yao, Gholami, Keutzer, and Mahoney]{yao2020pyhessian}
Yao, Z., Gholami, A., Keutzer, K., and Mahoney, M.~W.
\newblock Pyhessian: Neural networks through the lens of the hessian.
\newblock In \emph{2020 IEEE international conference on big data (Big data)},
  pp.\  581--590. IEEE, 2020.

\bibitem[Yin et~al.(2018)Yin, Pananjady, Lam, Papailiopoulos, Ramchandran, and
  Bartlett]{yin2018gradient}
Yin, D., Pananjady, A., Lam, M., Papailiopoulos, D., Ramchandran, K., and
  Bartlett, P.
\newblock Gradient diversity: a key ingredient for scalable distributed
  learning.
\newblock In \emph{International Conference on Artificial Intelligence and
  Statistics}, pp.\  1998--2007. PMLR, 2018.

\bibitem[Zhang et~al.(2018)Zhang, Wang, Xu, and Grosse]{zhang2018three}
Zhang, G., Wang, C., Xu, B., and Grosse, R.
\newblock Three mechanisms of weight decay regularization.
\newblock In \emph{International Conference on Learning Representations}, 2018.

\bibitem[Zielinski et~al.(2020)Zielinski, Krishnan, and
  Chatterjee]{zielinski2020weak}
Zielinski, P., Krishnan, S., and Chatterjee, S.
\newblock Weak and strong gradient directions: Explaining memorization,
  generalization, and hardness of examples at scale.
\newblock \emph{arXiv preprint arXiv:2003.07422}, 2020.

\bibitem[Zou \& Gu(2019)Zou and Gu]{zou2019improved}
Zou, D. and Gu, Q.
\newblock An improved analysis of training over-parameterized deep neural
  networks.
\newblock \emph{Advances in neural information processing systems}, 32, 2019.

\end{thebibliography}
\bibliographystyle{icml2023}

\newpage
\appendix
\onecolumn

\begin{center}
\Large
\textbf{Appendix}
\end{center}

In this appendix, we provide details omitted in the main paper and more experimental results and analyses.
\begin{itemize}
    \item \autoref{appendix:related_works}: more related works (cf. \autoref{sect:related_works} of the main paper).
    \item \autoref{appendix:more_results}: more experimental results and analyses (cf. \autoref{sect:gws}, \autoref{sect:hyperplane} and \autoref{sect:fedawo} of the main paper).
    \item \autoref{appendix:detail_fedawo}: additional details of \textsc{FedLAW} (cf. \autoref{sect:fedawo} of the main paper).
    \item \autoref{appendix:detail_implementation}: details of experimental setups (cf. \autoref{sect:gws}, \autoref{sect:hyperplane} and \autoref{sect:fedawo} of the main paper).
\end{itemize}

\section{More Related Works} \label{appendix:related_works}
\subsection{Model Aggregation in Federated Learning}
\textbf{Model aggregation in federated learning.} Model aggregation weights should be calibrated under asynchronous local updates. \textsc{FedNova} \citep{wang2020tackling} is proposed to tackle the objective inconsistency problem caused by asynchronous updates; it theoretically shows that the convergence will be improved if the numbers of local iterations normalize the aggregation weights. However, it does not take the heterogeneity degree of clients into account, which is also a key factor that affects the generalization of the global model. In \cite{DBLP:conf/iclr/ChenC21}, the authors point out that due to heterogeneity, the best-performing model will shift away from \textsc{FedAvg}, but they do not give insights on how to adjust aggregation weight to approximate the best model, they use Bayesian ensemble distillation method to prove the generalization of the global model instead. To solve the misalignment of neurons in FL with DNNs, \textsc{FedMA} \citep{wang2020federated} is proposed: \textsc{FedMA} constructs the shared global model layer-wise by matching and averaging hidden elements with similar features extraction signatures. Besides, optimal transport \citep{kantorovich2006translocation} can be adopted in layer-wise neuron alignment in the process of model fusion \citep{singh2020model}. These previous works improve the global model performance by layer-wise alignment, but they are complex and computation-expensive, and they can not be applied under the traditional weighted aggregation scheme. In this paper, we only focus on the convex combination of clients' local models by weighted aggregation, which is the most common and general way of model aggregation.

\subsection{Generalization and Training Dynamics of Neural Networks}
\textbf{Loss landscape of neural networks and generalization.} Deep neural networks (DNNs) are highly non-convex and over-parameterized, and visualizing the loss landscape of DNNs \citep{li2018visualizing, vlaar2021can} helps understand the training process and the properties of minima. There are mainly two lines of works about the loss landscape of DNNs. The first one is the linear interpolation of neural network loss landscape \citep{vlaar2021can,garipov2018loss,draxler2018essentially}, it plots linear slices of the landscape between two networks. In linear interpolation loss landscape, mode connectivity \citep{draxler2018essentially,vlaar2021can,entezari2021role} is referred to as the phenomenon that there might be increasing loss on the linear path between two minima found by SGD, and the loss increase on the path between two minima is referred to as (energy) barrier. It is also found that there may exist barriers between the initial model and the trained model \citep{vlaar2021can}. The second line concerns the loss landscape around a trained model's parameters \citep{li2018visualizing}. It is shown that the flatness of loss landscape curvature can reflect the generalization \citep{foret2020sharpness,izmailov2018averaging} and top hessian eigenvalues can present flatness \citep{yao2020pyhessian,jastrzkebski2018relation}. Networks with small top hessian eigenvalues have flat curvature and generalize well. Previous works seek flatter minima for improving generalization by implicitly regularizing the hessian \citep{foret2020sharpness,kwon2021asam,du2021efficient}. 

\textbf{Critical learning period in training neural networks.} \cite{jastrzebski2019break} found that the early phase of training of deep neural networks is critical for their final performance. They show that a break-even point exists on the learning trajectory, beyond which SGD implicitly regularizes the curvature of the loss surface and noise in the gradient. They also found that using a large learning rate in the initial phase of training reduces the variance of the gradient and improves generalization. In FL, \cite{yan2021critical} discovers the early training period is also critical to federated learning. They reduce the quantity of training data in the first couple of rounds and then recover the training data, and it is found that no matter how much data are added in the late period, the models still cannot reach a better accuracy. However, it did not further study the role of client heterogeneity in the critical learning period while we examine it by local gradient coherence.

\subsection{Federated Hyperparameter Optimization}
Current federated learning methods struggle in cases with heterogeneous client-side data distributions which can quickly lead to divergent local models and a collapse in performance. Careful hyperparameter tuning is particularly important in these cases. Hyperparameters can be optimized using gradient descent to minimize the final validation loss \citep{maclaurin2015gradient,franceschi2017forward}. Moreover, hyperparameters can be optimized based on reinforcement learning methods \citep{,guo2022auto,jomaa2019hyp,mostafa2019robust}. However, in this paper, optimizing aggregation weights is not our main novelty. Instead, we focus on leveraging this toolbox to examine the crucial training dynamics in FL in a principled way. 

\section{More Results and Analyses} \label{appendix:more_results}

\subsection{Global Weight Shrinking} \label{appendix:analysis_gws}
\textbf{Fixed $\gamma$.} We add more results about global weight shrinking experiments with fixed $\gamma$ as in \autoref{appdx_table:general_GWS}. It is found that when data are more NonIID, fixed $\gamma$ will cause negative effects; this is more dominant when $\alpha=0.1$ and the models are AlexNet or ResNet8.
\begin{table*} \small
\caption{\small More results about fixed $\gamma$ across different architectures in various NonIID settings.}
\centering
\begin{tabular}{c|cccccc}
\toprule
\textbf{$\gamma$}&\textbf{1.0} &\textbf{0.99} &\textbf{0.97} &\textbf{0.95} &\textbf{0.93} &\textbf{0.9} \\
\midrule 
Model &\multicolumn{6}{c}{$\alpha=10$}\\
\midrule
SimpleCNN&65.96	&67.19	&69.41	&\textbf{69.81}	&69.69	&69.59\\
AlexNet&73.9	&74.43	&74.96	&75.12	&\textbf{75.33}	&74.06\\
ResNet8&76.26	&75.63	&76.92	&\textbf{77.23}	&76.9	&76.61\\
\midrule
Model &\multicolumn{6}{c}{$\alpha=0.5$}\\
\midrule
SimpleCNN&65.78	&66.59	&67.93	&\textbf{68.85}	&68.75	&68.25\\
AlexNet&73.72	&73.06	&73.89	&\textbf{73.98}	&73.6	&73.33\\
ResNet8&73.4	&73.93	&\textbf{75.39}	&74.12	&73.66	&73.46\\
\midrule
Model &\multicolumn{6}{c}{$\alpha=0.2$}\\
\midrule
SimpleCNN&63.52	&64.68	&63.72	&\textbf{65.82}	&65.4	&64.97\\
AlexNet&68.41	&70.46	&\textbf{70.87}	&70.74	&70.58	&69.42\\
ResNet8&71.85	&70.96	&\textbf{72.76}	&72.04	&71.25	&62.32\\
\midrule
Model &\multicolumn{6}{c}{$\alpha=0.1$}\\
\midrule
SimpleCNN&60.57	&61.22	&61.83	&\textbf{62.05}	&62.05	&60.85\\
AlexNet&\textbf{66.18}	&65.25	&64.74	&64.23	&64.16	&61.24\\
ResNet8&\textbf{63.89}	&60.55	&61.38	&59.23	&58.76	&39.85\\
\bottomrule
\end{tabular}
\label{appdx_table:general_GWS}
\end{table*}

\begin{table*}[]
\centering
\caption{The performance of adaptive GWS under different global learning rates.}
\begin{tabular}{c|ccc|ccc}
\hline
                     & \multicolumn{3}{c|}{IID ($\alpha=100$)} & \multicolumn{3}{c}{NonIID ($\alpha=1$)} \\ \hline
Global learning rate & 0.5       & 1         & 1.5      & 0.5    & 1      & 1.5            \\ \hline
FedAvg               & 69.15     & 68.18     & 64.35    & 68.71  & 67.09  & 64.00          \\
Adaptive GWS         & 71.45     & 71.98     & 71.13    & 69.65  & 71.02  & 71.04 \\
$\gamma$ of Adaptive GWS    & 0.986     & 0.974     & 0.963    & 0.991  & 0.979  & 0.967          \\ \hline
\end{tabular}
\label{tab:gws_global_lr}
\end{table*}

\textbf{Adaptive GWS with global learning rate.} We conduct experiments with the adaptive GWS under different global learning rates for both IID and NonIID settings. We train SimpleCNN on CIFAR10 with 1 local epoch, and the results are reported in \autoref{tab:gws_global_lr}. It can be observed that in both IID and NonIID settings, a small global server learning rate can improve \textsc{FedAvg}'s performance. In contrast, the larger the global learning rate, the smaller the learned $\gamma$ (stronger regularization). It is aligned with our insights in the main paper that larger pseudo gradients require stronger regularization. Moreover, adaptive GWS is robust to the choice of the global server learning rate, especially in the IID setting.

\begin{figure*}[h]
\centering
\includegraphics[width=0.40\columnwidth]{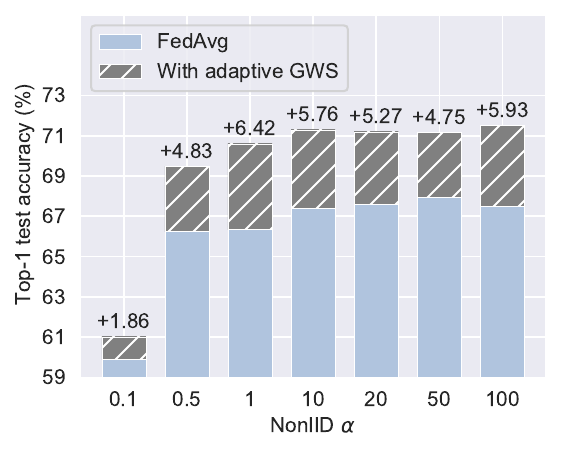}
\includegraphics[width=0.40\columnwidth]{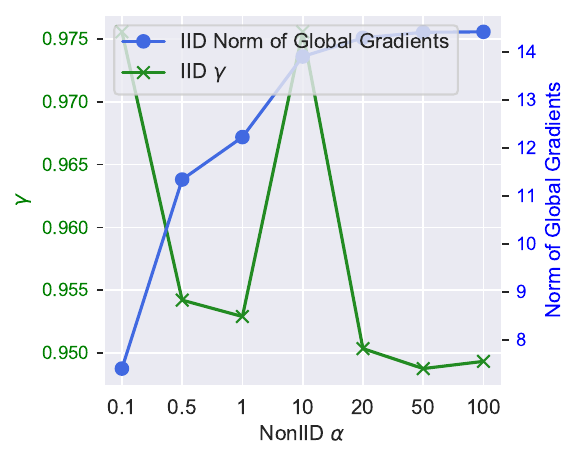}
\caption{\small \textbf{Adaptive GWS under various heterogeneity.}  \textbf{Left:} Test accuracy gains with adaptive GWS. In all settings, adaptive GWS can bring performance gains. \textbf{Right:} Learned $\gamma$ of adaptive GWS in different settings. $\gamma$ decreases when data become more IID, causing the stronger weight shrinking effect. This is due to the changes in the volumes of global gradients. The norm of global gradient increases when data become more IID, and it requires smaller $\gamma$ to cause stronger regularization.}
\label{appdx_fig:gws_testacc_gamma}
\end{figure*}

\textbf{Adaptive GWS under various heterogeneity.} We show adaptive GWS works under various heterogeneity and visualize $\gamma$ and the norm of the global gradient in each setting, as in \autoref{appdx_fig:gws_testacc_gamma}. It demonstrates that adaptive GWS can boost performance under different NonIID settings, but it has a smaller benefit when the system is extremely NonIID (i.e., $\alpha=0.1$). Additionally, according to the right figure of \autoref{appdx_fig:gws_testacc_gamma}, except for the outlier $\gamma$ when $\alpha=10$, the learned $\gamma$ decreases when data become more IID, causing stronger weight shrinking effect. We think this is a result of a balance between optimization and regularization. The volumes of global gradients change when the heterogeneity changes. The norm of global gradient increases when data become more IID, and it requires smaller $\gamma$ to cause stronger regularization.

\begin{figure*}[h]
\centering
\includegraphics[width=0.32\columnwidth]{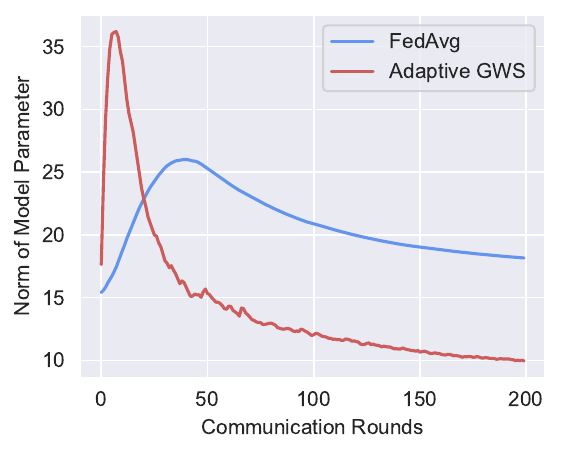}
\includegraphics[width=0.32\columnwidth]{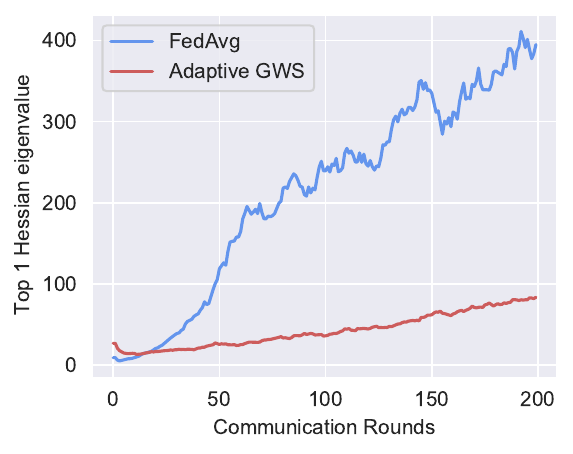}
\includegraphics[width=0.32\columnwidth]{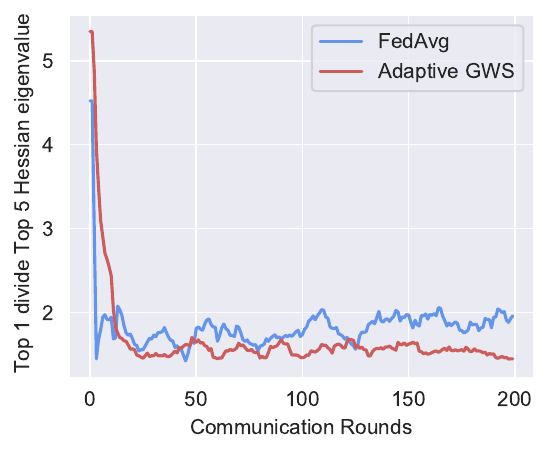}
\caption{\small \textbf{More results of general understanding of adaptive GWS.} \textbf{Left:} Adaptive GWS results in a smaller model parameter during training. \textbf{Middle:} Smaller top 1 hessian eigenvalue indicates flatter curvature of DNNs. The result shows \textsc{FedAvg} tends to generate sharper global models during training while adaptive GWS seeks flatter networks. \textbf{Right:} The ratio of the top 1 hessian eigenvalue and top 5 is another indicator; a smaller value means flatter minima.}
\label{appdx:fig_general_understand}
\end{figure*}

\textbf{More results of general understanding of adaptive GWS.} First, we first visualize the norm of model parameter weight during training as in the left figure of \autoref{appdx:fig_general_understand}. Adaptive GWS results in a smaller model parameter during training. Second, we use two common metrics to measure the flatness of loss landscape during training as in the middle and right figures of \autoref{appdx:fig_general_understand}, and they are the hessian eigenvalue-based metrics. The dominant hessian eigenvalue evaluates the worst-case loss landscape, which means the larger top 1 eigenvalue indicates the greater change in the loss along this direction and the sharper the minima \citep{keskar2017large}. We adopt the top 1 hessian eigenvalue and the ratio of top 1 and top 5, which are commonly used as a proxy for flatness \citep{jastrzebski2020break,fort2019large}. Usually, a smaller top 1 hessian eigenvalue and a smaller ratio of top 1 hessian eigenvalue and top 5 indicates flatter curvature of DNN. As in the figures, during the training, \textsc{FedAvg} generates global models with sharp landscapes whereas adaptive GWS tends to generate more generalized models with flatter curvatures. 

\begin{figure*}[h]
\centering
\includegraphics[width=0.6\columnwidth]{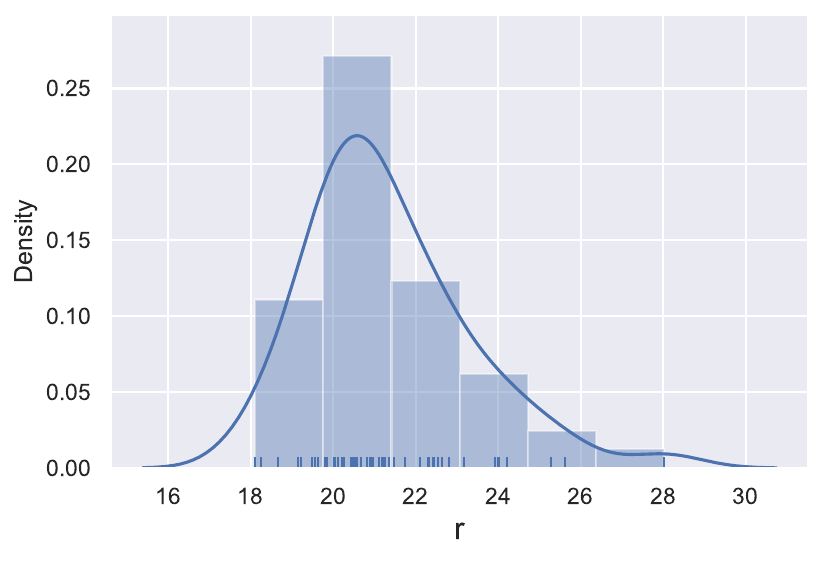}
\caption{\small \textbf{Distribution of $r$.} We visualize $r$ values of all experiments in \autoref{fig:fix_gamma_test_gain} and \autoref{appdx_fig:gws_testacc_gamma} and find that the distribution of $r$ can be approximated into a Gaussian distribution with its mean around 20.5.}
\label{appdx_fig:distr_ratio}
\end{figure*}

\textbf{The distribution of $r$.} We visualize $r$ (the ratio of the global gradient and the regularization pseudo gradient) values of all experiments in \autoref{fig:fix_gamma_test_gain} and \autoref{appdx_fig:gws_testacc_gamma} as in \autoref{appdx_fig:distr_ratio}. It is found that the distribution of $r$ can be approximated into a Gaussian distribution with its mean around 20.5.

\subsection{Client Coherence} \label{appendix:analysis_client_coherence}
\textbf{The relationship with gradient diversity.} 
The conclusion of gradient diversity \citep{yin2018gradient} is opposite to the one of gradient coherence. Gradient diversity argues that higher similarities between workers' gradients will degrade performance in distributed mini-batch SGD, while gradient coherence claims that higher similarities between the gradients of samples will boost generalization \citep{yin2018gradient,chatterjee2019coherent}. Moreover, gradient diversity is somewhat controversial. As argued in the line of works about gradient coherence \citep{chatterjee2020making,chatterjee2019coherent}, the manuscript of gradient diversity did not explicitly measure the gradient diversity in the experiments (or further study its properties): only experiments on CIFAR-10 can be found where they replicate $1/r$ of the dataset $r$ times and show that greater the value of r less the effectiveness of mini-batching to speed up. Apart from this controversy, the strongly-convex assumption in the theorem of gradient diversity \citep{yin2018gradient} may make it weaker to generalize its conclusions in neural networks while we are studying the empirical properties in FL with neural networks. Taking the above statements into consideration, gradient diversity may be infeasible in our settings.

\textbf{The relationship with client similarity works in FL.} 
There are some works \citep{karimireddy2020scaffold,li2020federated} taking the bounded gradient dissimilarity assumption to deduce theorems. In their assumptions, they bound the gradient sum or gradient norm, but we use the cosine similarity to study how the clients interplay with each other and contribute to the global. So the perspectives are quite different. Additionally, there are previous works in FL that use cosine similarity of clients’ gradients to improve personalization \cite{huang2021personalized,li2022towards}; however, we focus on the training dynamics in generalization, and one of our novel findings is we discover a critical point exists and the periods that before or after this point play different roles in the global generalization. 

\begin{figure*}[h]
\centering
\includegraphics[width=0.28\columnwidth]{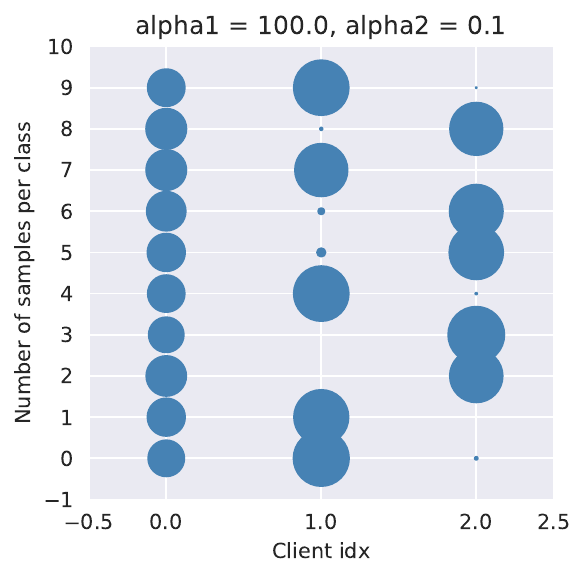} 
\includegraphics[width=0.35\columnwidth]{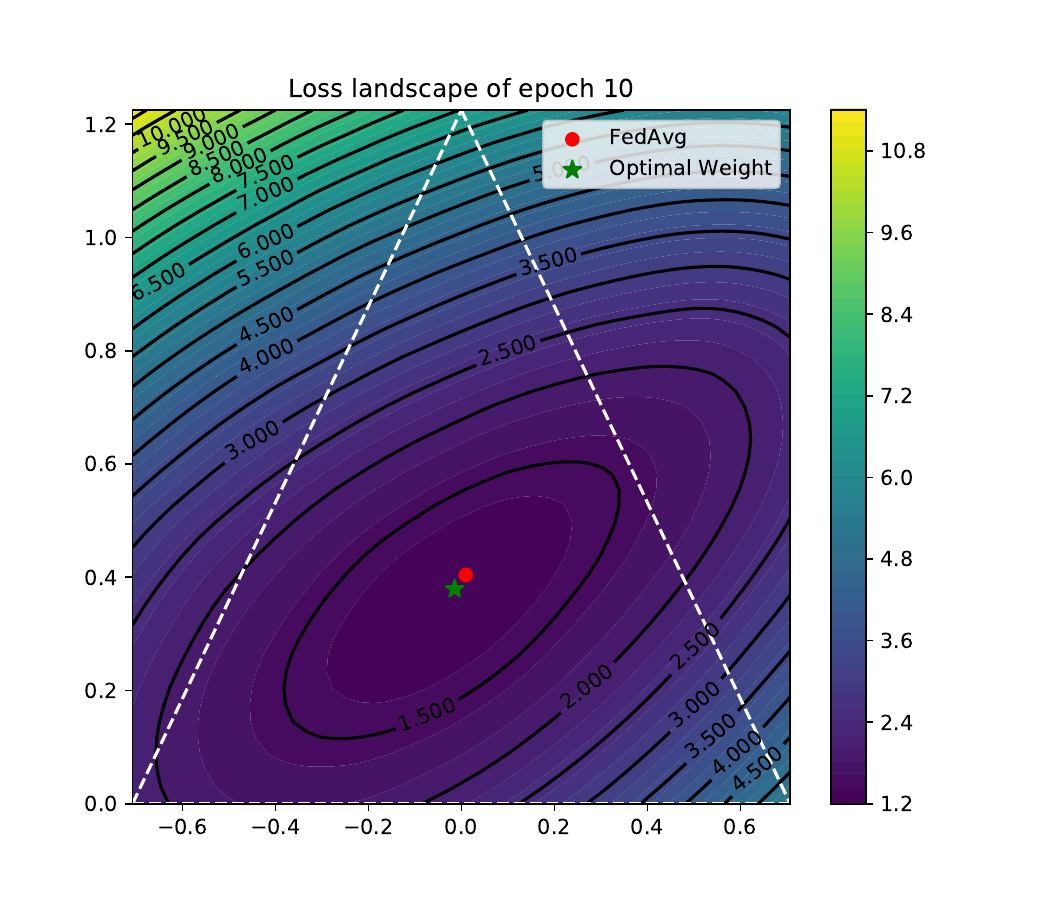} 
\includegraphics[width=0.35\columnwidth]{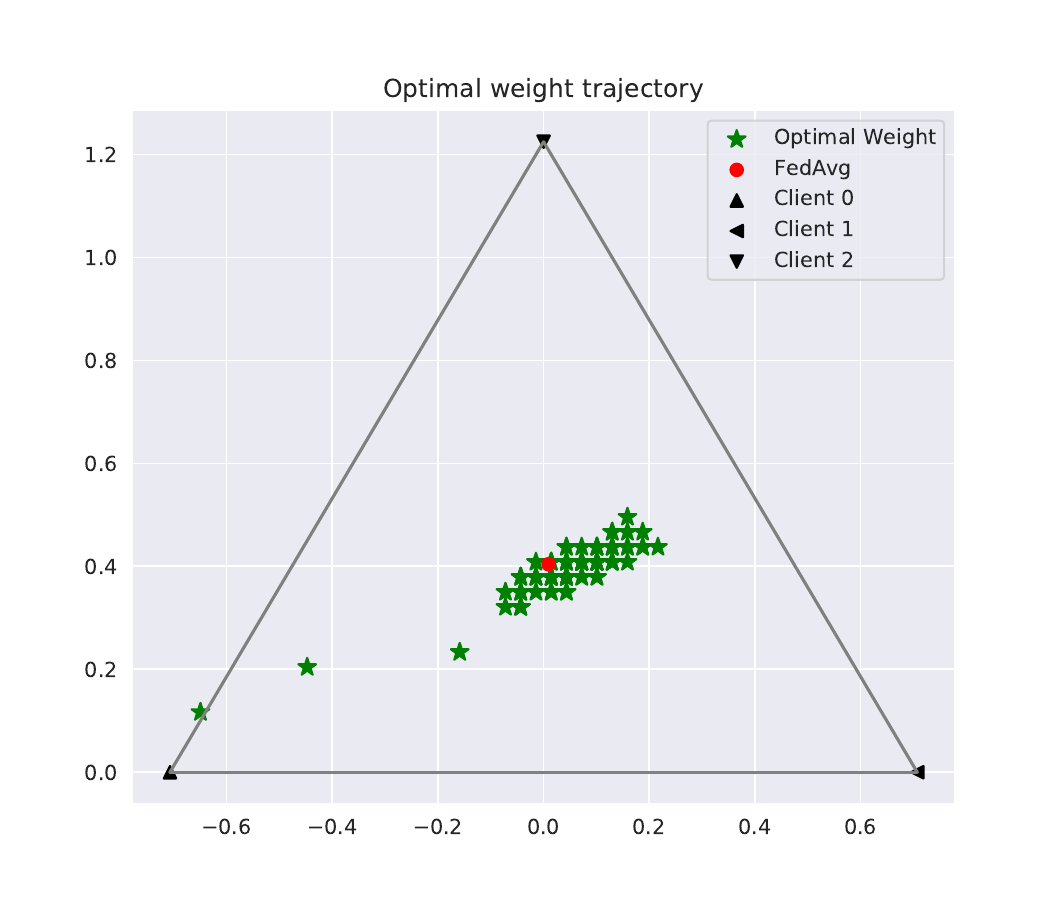} 
\includegraphics[width=0.28\columnwidth]{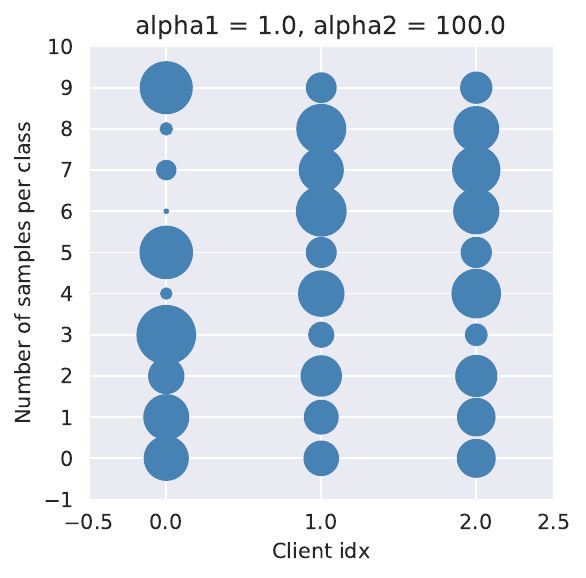} 
\includegraphics[width=0.35\columnwidth]{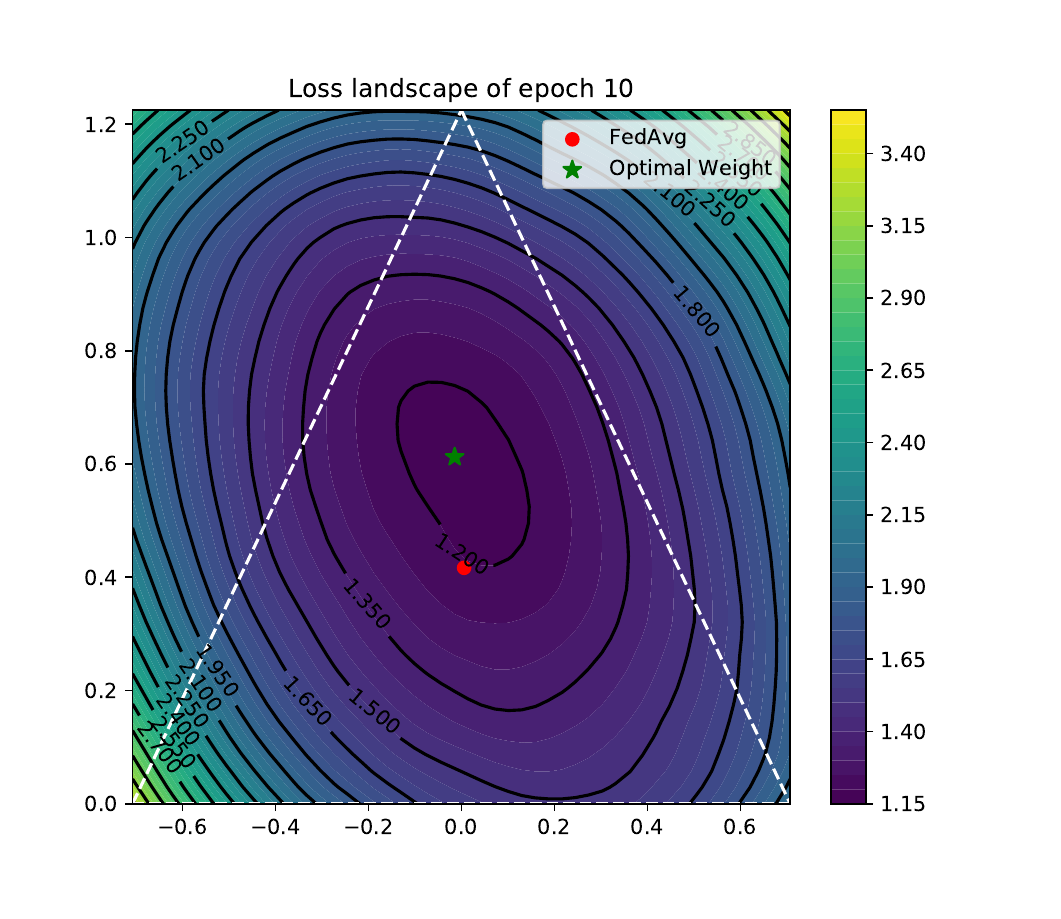} 
\includegraphics[width=0.35\columnwidth]{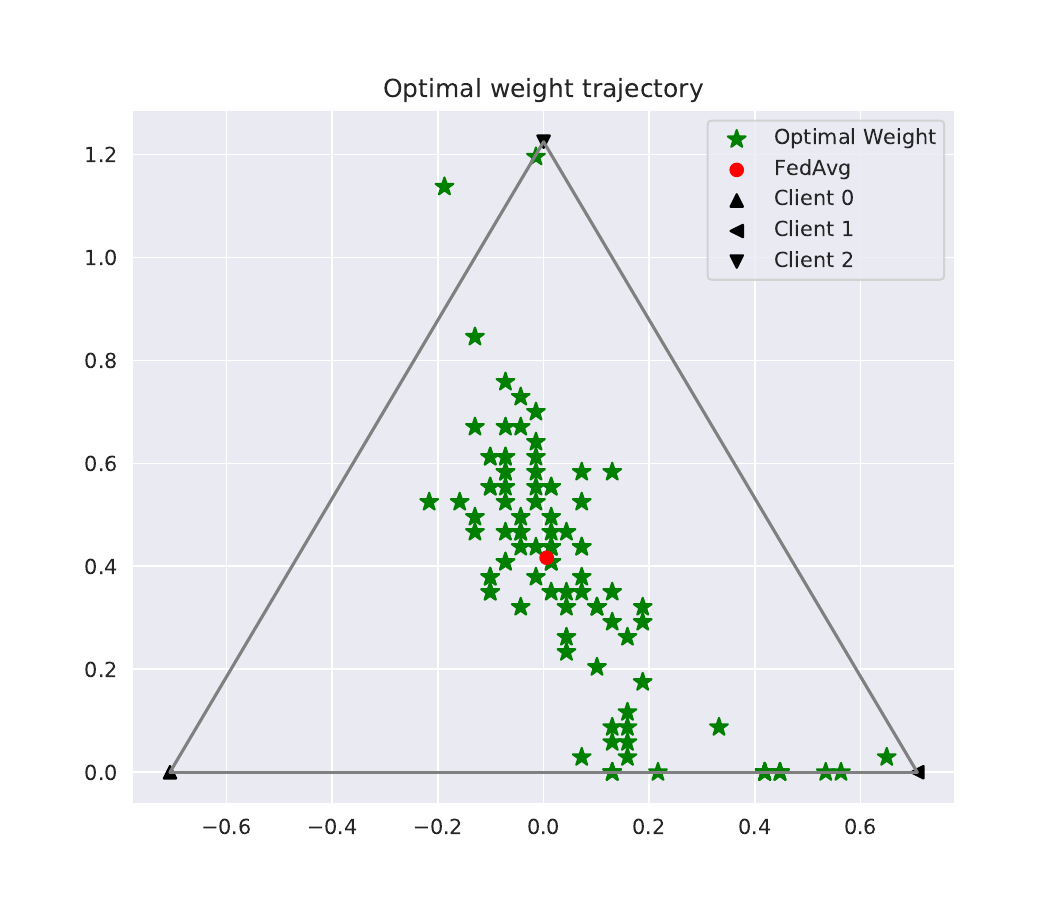} 
\includegraphics[width=0.28\columnwidth]{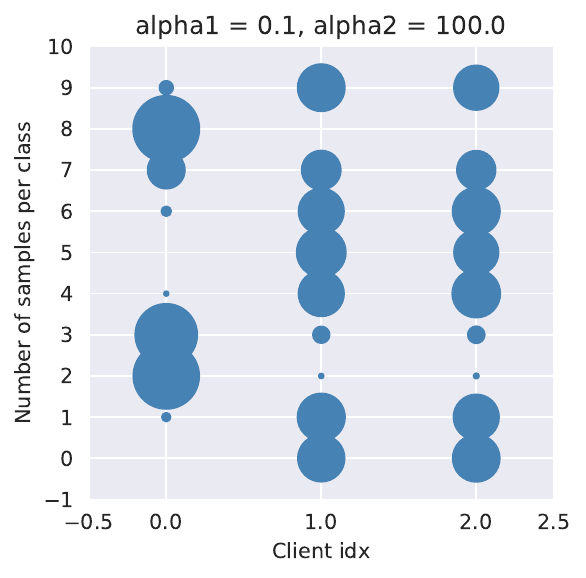}
\includegraphics[width=0.35\columnwidth]{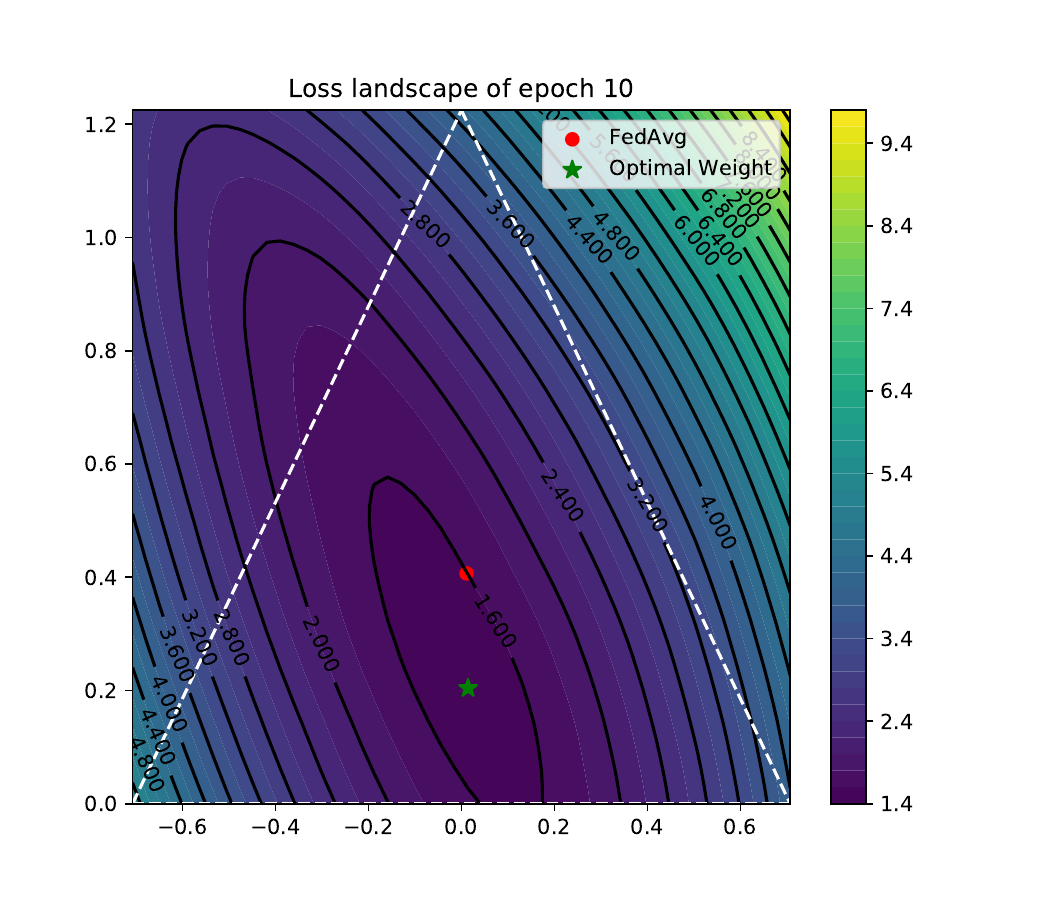}
\includegraphics[width=0.35\columnwidth]{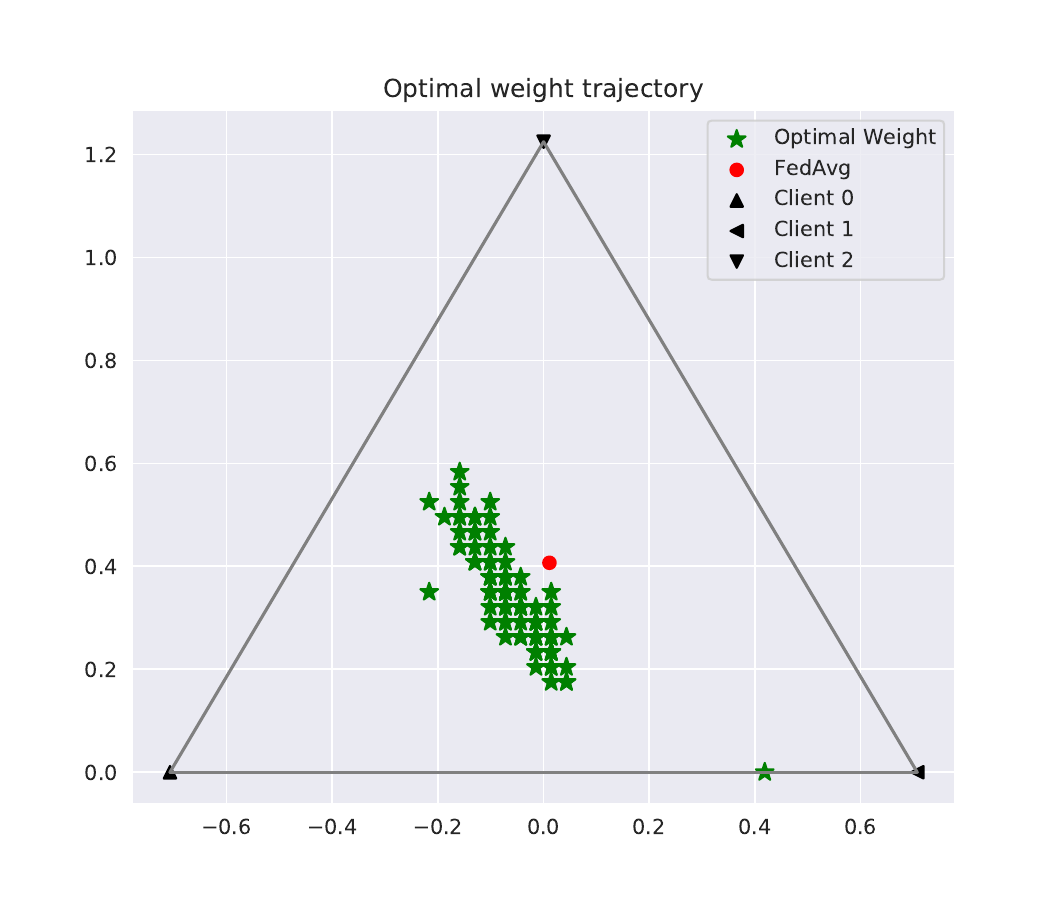}
\caption{\textbf{Heterogeneity also affects the optimal aggregation weight.} A three-node toy example on CIFAR-10 is shown. \textbf{Left:} Data distribution of each client, note that each client has the same dataset size. \textbf{Middle:} Loss landscape on the plane of aggregation weight, it is noticed that \textsc{FedAvg} is off the optimal and the landscape has various flatness in different directions. \textbf{Right:} optimal weight trajectory during training. We plot the optimal weights in each round (green dots) and find that the optimal weights are biased from \textsc{FedAvg}.}
\label{appdx_fig:toy_example}
\end{figure*}

\textbf{Visualization of how heterogeneity affects the optimal aggregation weight.} We set up a three-node toy example on CIFAR-10 by hybrid Dirichlet sampling as shown in \autoref{appdx_fig:toy_example}. We first sample client 0's data distribution by Dirichlet sampling according to $\alpha_{1}$; then we sample data distributions for clients 1 and 2 on the remaining data with $\alpha_{2}$. We set up three settings with different $\alpha_{1}, \alpha_{2}$ and illustrate the data distributions on the \textbf{Left column} in \autoref{appdx_fig:toy_example}. In the example, the aggregation weights (AWs) are $[\lambda_{0}, \lambda_{1}, \lambda_{2}]$, we regularize the weights as $\lambda_{0} + \lambda_{1} + \lambda_{2} = 1$ which is a plane that can be visualized in 2-D. We uniformly sample points on the plane to obtain global models with different AW and compute the test loss, and then the loss landscapes on the plane can be visualized. We implement \textsc{FedAvg} for 100 rounds and record the loss landscape and the optimal weight on the loss landscape in each round; then we illustrate the loss landscape of round 10 on the \textbf{Middle column} and the optimal weight trajectory on the \textbf{Right column} of \autoref{appdx_fig:toy_example}. 

In these settings, clients have different heterogeneity degrees: in the first setting, client 0 has a balanced dataset while the data of clients 1 and 2 are complementary; in the second and third settings, clients 1 and 2 have the same data distribution, which differs from the client 0's. From \autoref{appdx_fig:toy_example}, it is evident that the weight of \textsc{FedAvg} is biased from optimal weights when heterogeneity degrees vary in clients, we can draw the following conclusions: (1) optimal weight can be viewed as a Gaussian distribution in the aggregation weight hyperplane; (2) the mean of the Gaussian will drift towards to the directions where data are more inter-heterogeneous (for instance, in the third setting, client 0's major classes are 2, 3 and 8 while client 1 and 2 have rare data on these classes, so client 0's contribution is more dominant); (3) the variance of the Gaussian is larger in inter-homogeneous direction and is smaller in inter-heterogeneous direction (the variance along the client 1-client 2 direction is large in the second and third settings, because the two clients have inter-homogeneous data; opposite phenomenon is shown in the first setting, where client 1 and 2 have inter-heterogeneous data); (4) the flatness of loss landscape on aggregation weight hyperplane is consistent with the variance of the Gaussian, which means the directions with more significant variance will have flatter curvature in the landscape. From our analysis, it is clear that clients' contributions to the global model should not be solely measured by dataset size, and the heterogeneity degree should also be taken into account. And we observe that in a more heterogeneous environment, the loss landscape is sharper, which means the bias from optimal weight will cause more generalization drop. In other words, in a heterogeneous environment, appropriate aggregation weight matters more.

\begin{figure*}[h]
\centering
\includegraphics[width=0.7\columnwidth]{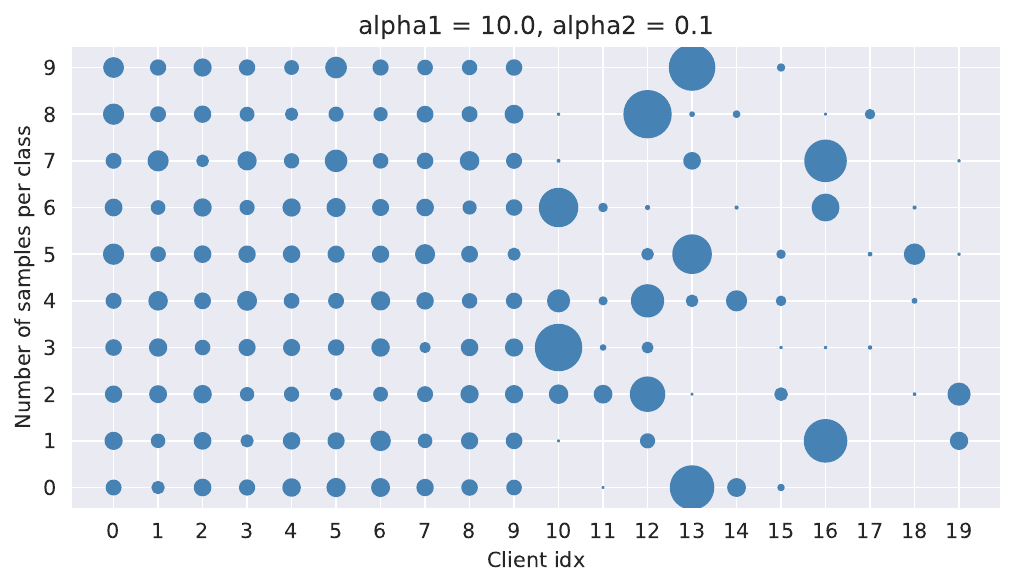}
\caption{\small \textbf{Data distribution of \autoref{fig:hyperplane_noniid}}}
\label{appdx_fig:hyperplane_noniid_distr}
\end{figure*}

\textbf{Visualization of the hybrid NonIID setting of \autoref{fig:hyperplane_noniid}.} We visualize the hybrid NonIID setting of \autoref{fig:hyperplane_noniid} in \autoref{appdx_fig:hyperplane_noniid_distr}. We take $\alpha_1 = 10$ and $\alpha_2=0.1$, so the first 10 clients (indexed 0-9) have class-balanced data while the last 10 clients (indexed 10-19) have class-imbalanced data.

\begin{table}[h] \small
\centering
\caption{\textbf{Pearson correlation coefficient analysis of AW.} Heterogeneity degree is calculated as the reciprocal of the variance of class distribution for each client. We take the accumulated weights during the training as clients' AW.}
\begin{tabular}{c|ccc}
\toprule
Factors&$E=1$&$E=5$&$E=1\&E=5$\\
\midrule
Dataset size (DS)&-0.098 &\textbf{0.21} &0.035\\
Heterogeneity degree (HD)&\textbf{0.41} &0.024 &0.35\\
DS$\times$HD&0.26 &0.17 &\textbf{0.31}\\
\bottomrule
\end{tabular}
\label{table:pearson}
\end{table}
\textbf{Data size or heterogeneity? A correlation analysis.} Data size and heterogeneity all affect clients' contributions to the global model, but which affects it most? As in previous literature, the importance is depicted by the dataset size that clients with more data will be assigned larger weights. According to the analysis in \autoref{appdx_fig:toy_example}, the importance of weight may be associated with the heterogeneity degrees of clients. To explore which factor is more dominant in the AW optimized by attentive LAW, we have made a Pearson correlation coefficient analysis in \autoref{table:pearson}. Results show that dataset size is more dominant when the local epoch is large; otherwise, the heterogeneity degree. This phenomenon is intuitive: when the local epoch increases, clients with a larger dataset will have more local iterations than others \citep{wang2020tackling}, so their updates are more dominant. In the cases where the local epoch is small, clients' updates are of similar volumes; here the updates' directions are much more important since balanced clients are prone to have stronger coherence, and their AWs are larger in model aggregation. We combine two factors by multiplication, and the result shows that the combined indicator is more dominant when the two cases are mixed.

\subsection{Learning curves of FedLAW and baselines}
\begin{figure}[h]
\centering
\includegraphics[width=0.49\columnwidth]{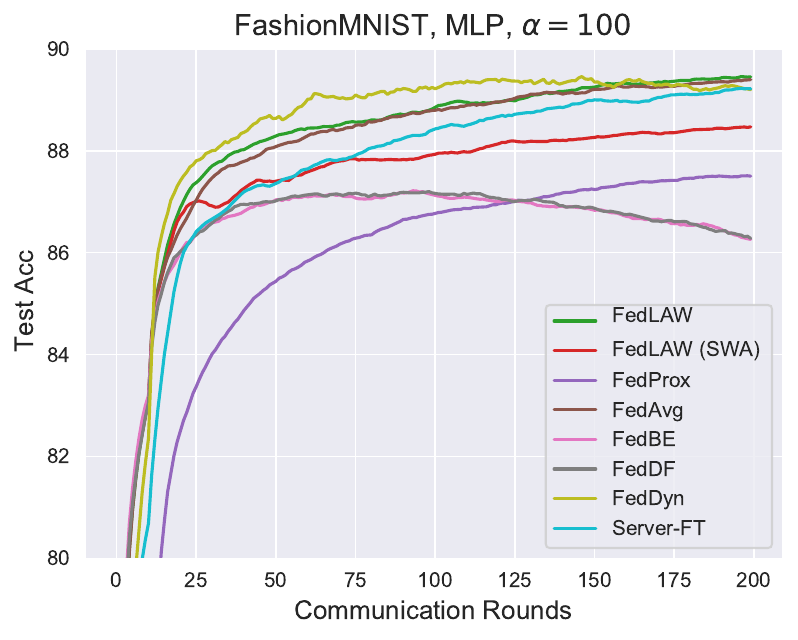}
\includegraphics[width=0.49\columnwidth]{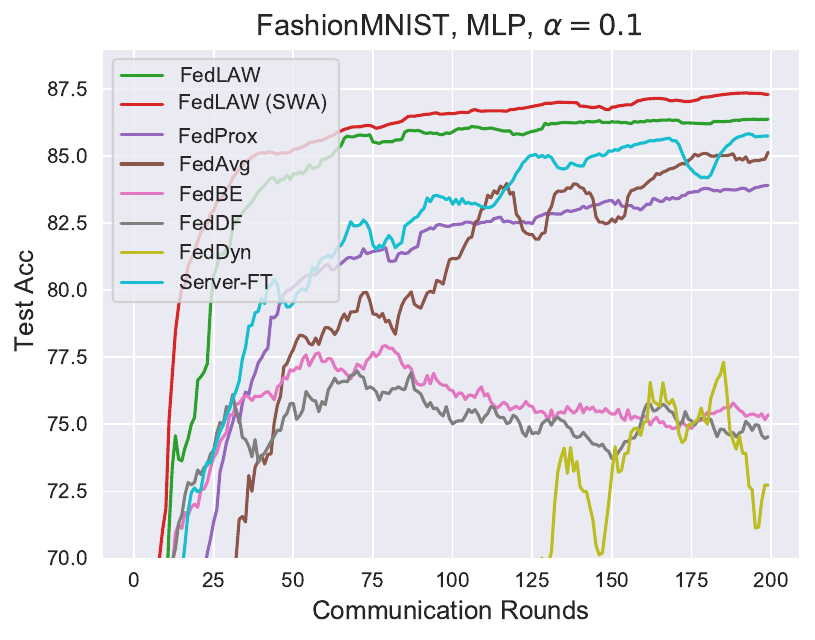}
\includegraphics[width=0.49\columnwidth]{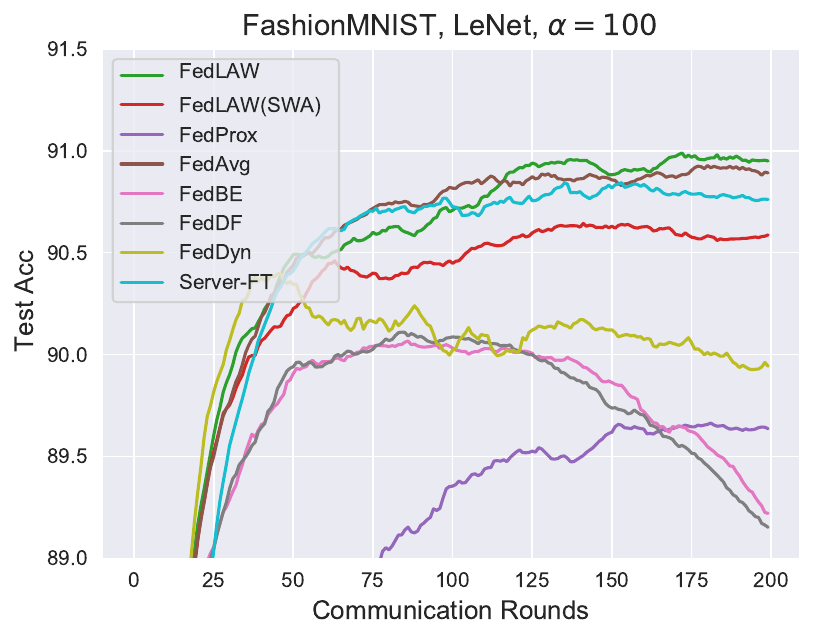}
\includegraphics[width=0.49\columnwidth]{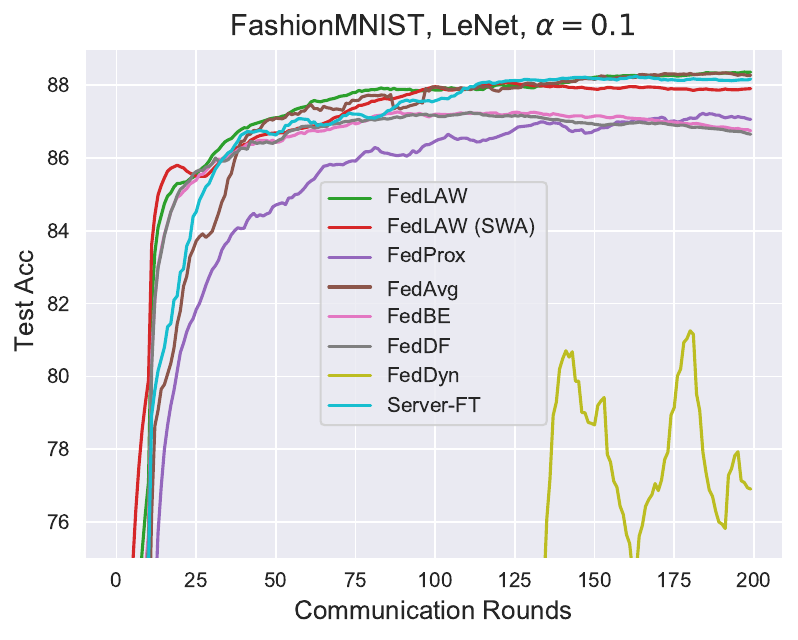}
\caption{\small \textbf{Test accuracy curves of algorithms under FashionMNIST.} According to the results in \autoref{table:first_table}. }
\label{appdx_fig:curve_fmnist}
\vspace{-0.5cm}
\end{figure}

\begin{figure}[h]
\centering
\includegraphics[width=0.49\columnwidth]{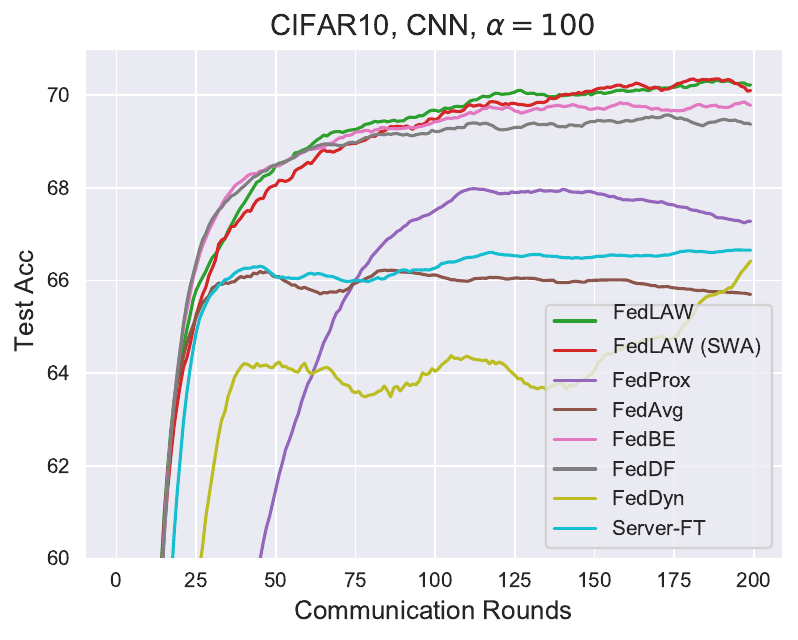}
\includegraphics[width=0.49\columnwidth]{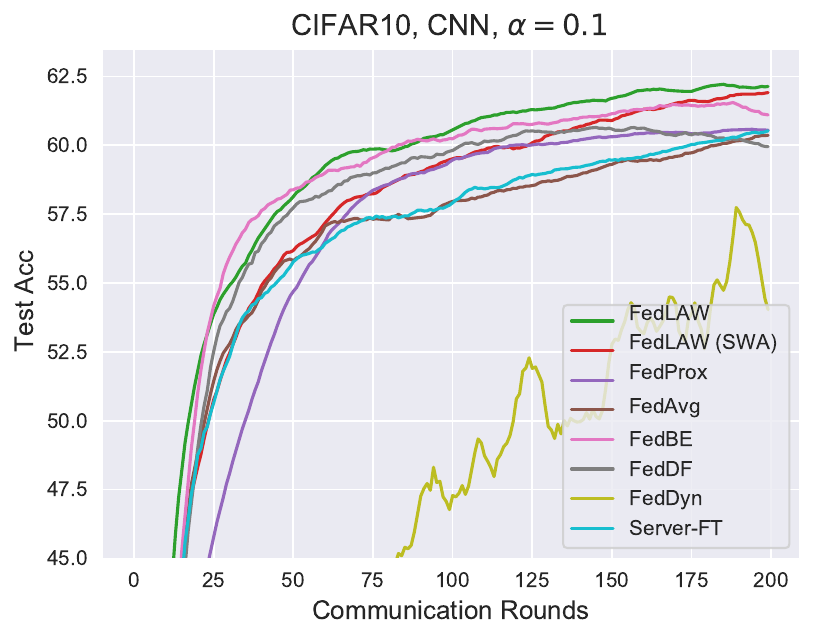}
\includegraphics[width=0.49\columnwidth]{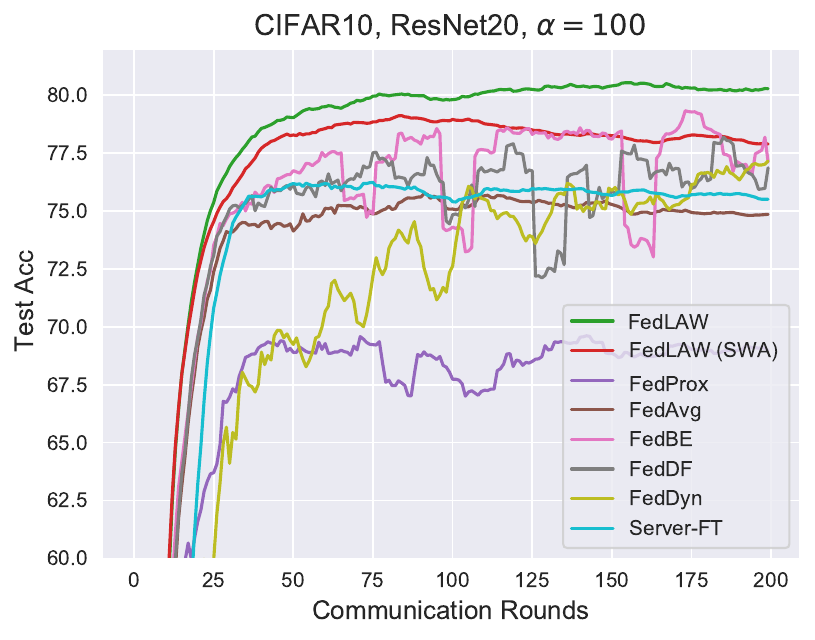}
\includegraphics[width=0.49\columnwidth]{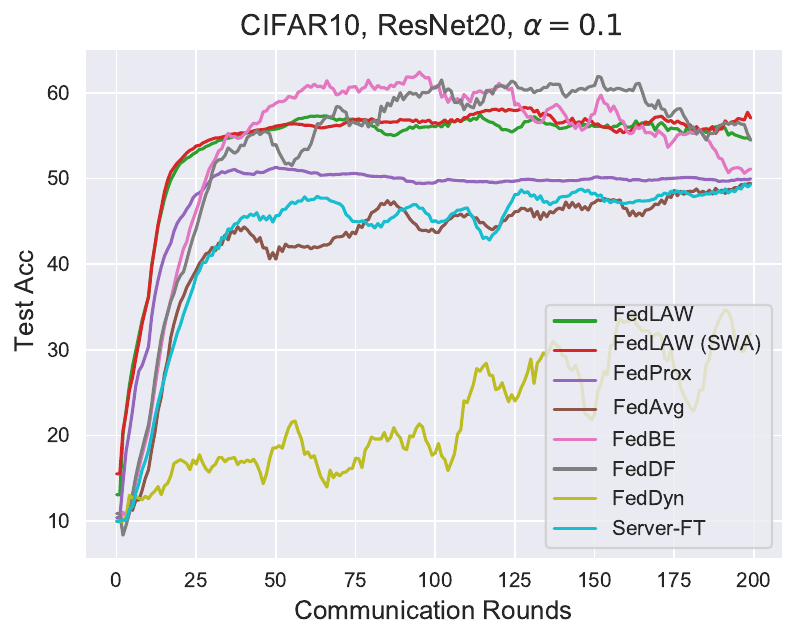}
\caption{\small \textbf{Test accuracy curves of algorithms under CIFAR-10.} According to the results in \autoref{table:first_table}. }
\label{appdx_fig:curve_CIFAR-10}
\vspace{-0.5cm}
\end{figure}

\begin{figure}[h]
\centering
\includegraphics[width=0.49\columnwidth]{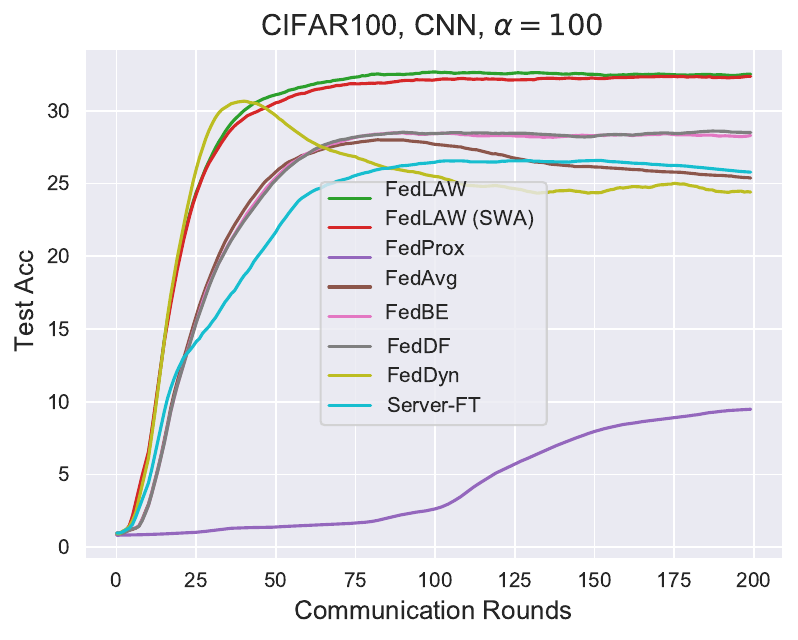}
\includegraphics[width=0.49\columnwidth]{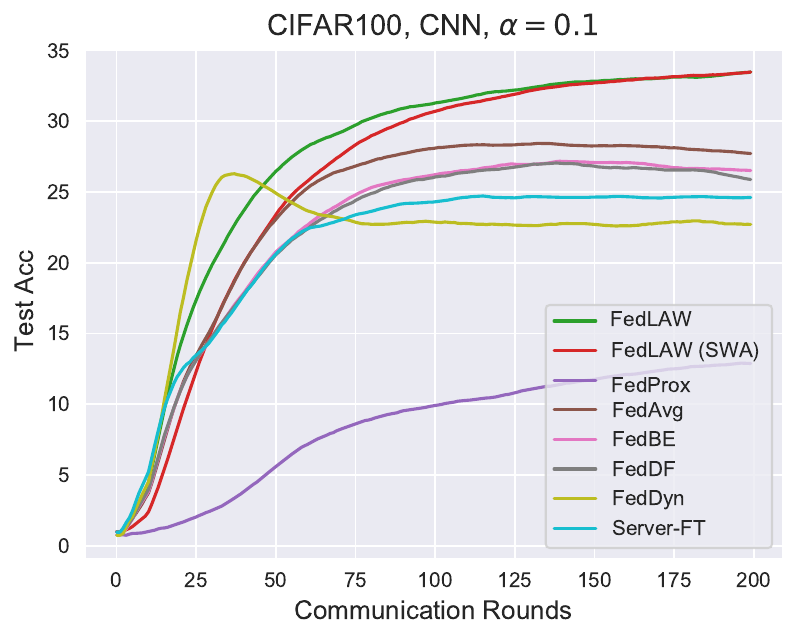}
\includegraphics[width=0.49\columnwidth]{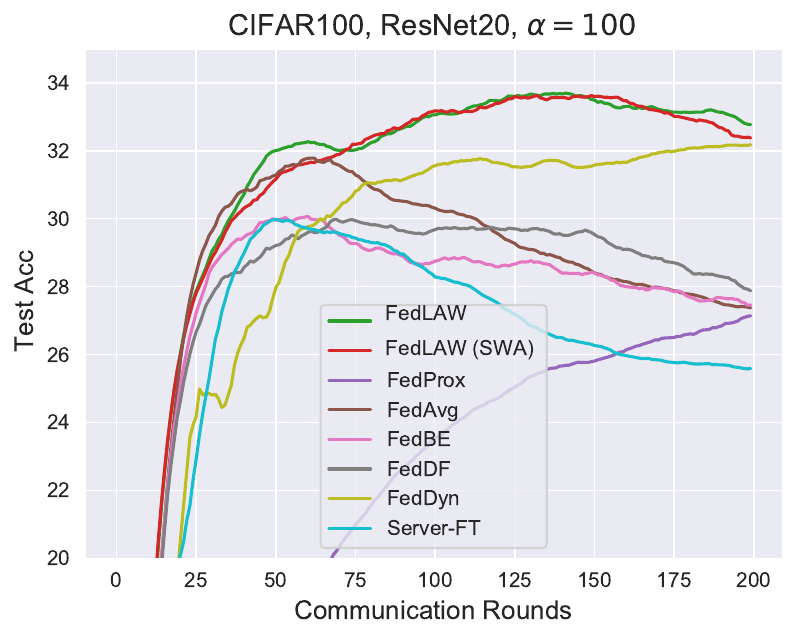}
\includegraphics[width=0.49\columnwidth]{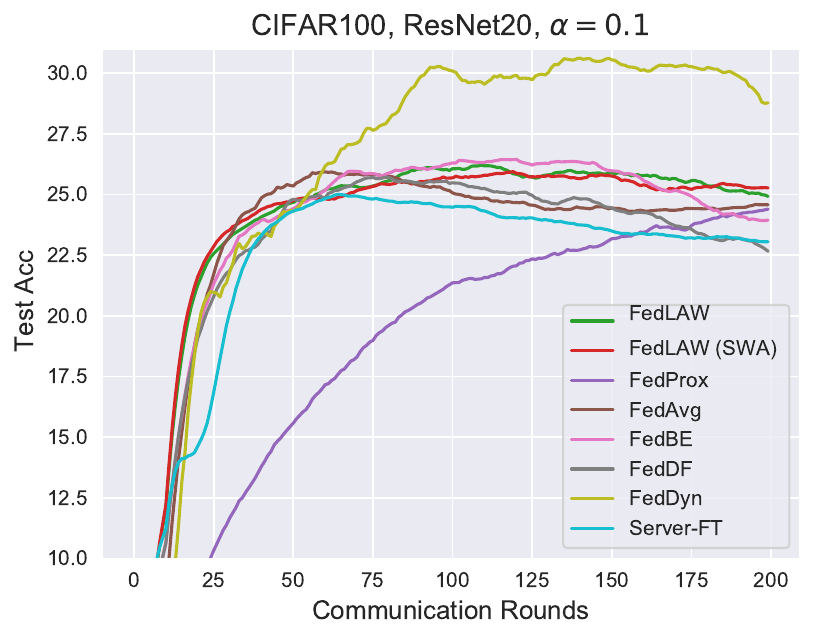}
\caption{\small \textbf{Test accuracy curves of algorithms under CIFAR-100.} According to the results in \autoref{table:first_table}. }
\label{appdx_fig:curve_CIFAR-100}
\vspace{-0.5cm}
\end{figure}

We add the test accuracy curves to show the learning processes of the algorithms and visualize them in \autoref{appdx_fig:curve_fmnist} (FasionMNIST), \autoref{appdx_fig:curve_CIFAR-10} (CIFAR-10), and \autoref{appdx_fig:curve_CIFAR-100} (CIFAR-100). The curves are according to the results in \autoref{table:first_table}. It shows that \textsc{FedLAW} surpasses the baseline algorithms in most cases. Besides, \textsc{FedLAW} is steady in the learning curves and it avoids over-fitting in the late training. 

\begin{figure*}[h]
\centering
\includegraphics[width=0.3\columnwidth]{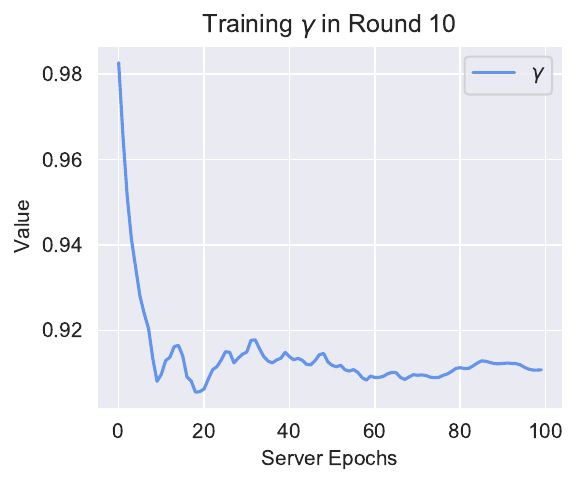}
\includegraphics[width=0.3\columnwidth]{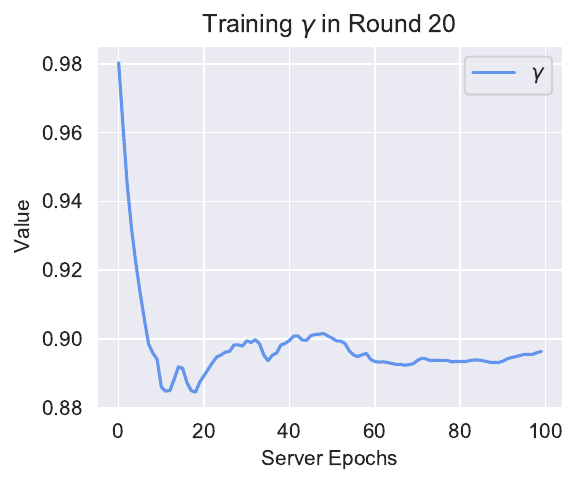}
\includegraphics[width=0.3\columnwidth]{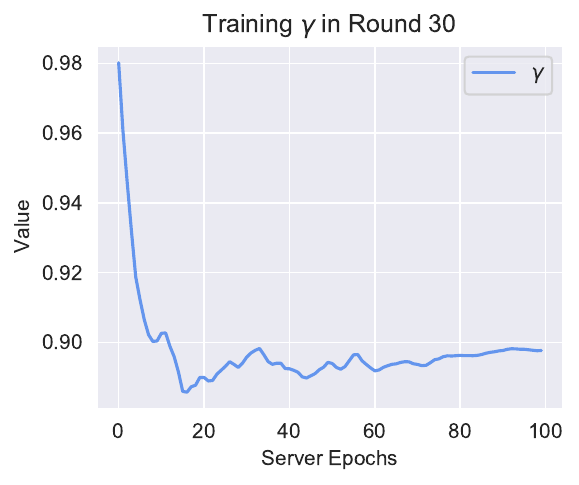}
\includegraphics[width=0.3\columnwidth]{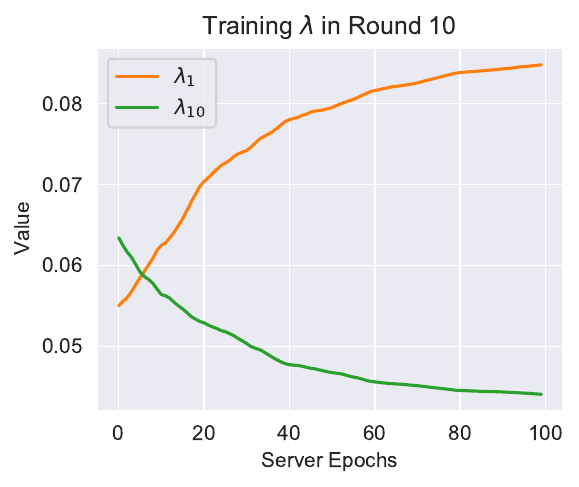}
\includegraphics[width=0.3\columnwidth]{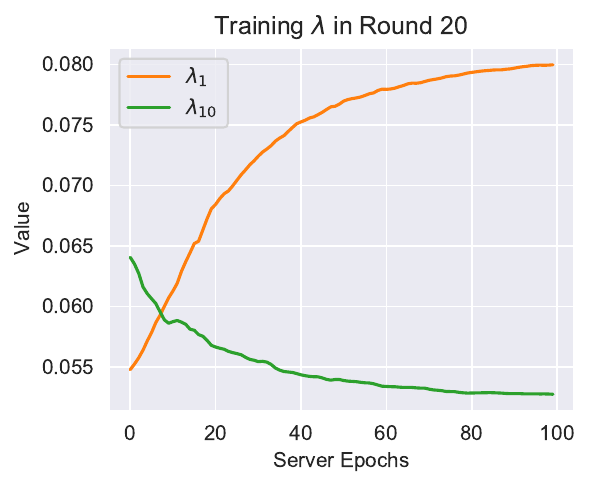}
\includegraphics[width=0.3\columnwidth]{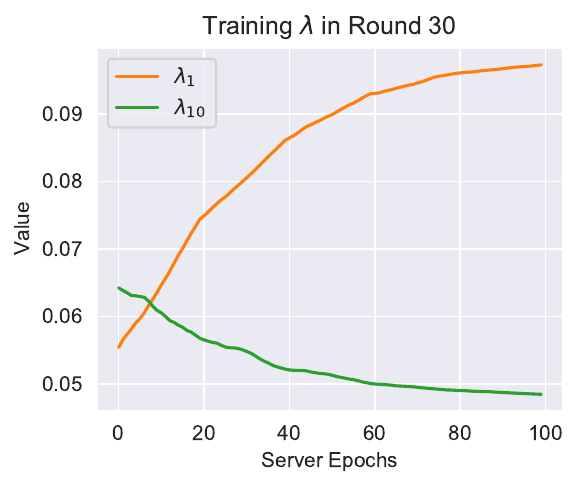}
\caption{\small \textbf{Server training visualization of \textsc{FedLAW}.} CIFAR10, $n=20$, $E=3$, NonIID $\alpha=1.0$, ResNet20. }
\label{appdx_fig:server_training_vis}
\end{figure*}

We also visualize the server training process of FedLAW in \autoref{appdx_fig:server_training_vis}. It is found that $\gamma$ converges faster than $\lambda$. For $\gamma$, it converges to the optimal value in about 30 server epochs, while for $\lambda$, it needs 80 epochs to fully converge.

\section{Additional Details of FedLAW} \label{appendix:detail_fedawo}
In \textsc{FedLAW}, we optimize AW on the server as \autoref{equ:fedawo}, and there are constraints that $\lambda_{i} \geq 0, \Vert \boldsymbol{\lambda} \Vert_{1} = 1$. To realize these constraints, we adopt base functions in $\lambda$, and there are two alternatives, the quadratic function and the exponential function.
\begin{equation}
    \text{Quadratic: }\lambda_{i} = \frac{x_{i}^{2}}{\sum_{j}^{m} x_{j}^{2}}; \text{Exponential: }\lambda_{i} = \frac{e^{x_{i}}}{\sum_{j}^{m} e^{x_{j}}}.
\end{equation}
$\boldsymbol{x}$ is the variable that determines the value of $\boldsymbol{\lambda}$. We compute the gradients of $\boldsymbol{x}$ to update $\boldsymbol{\lambda}$. By using the base functions, $\boldsymbol{\lambda}$ can meet the constraints of non-negativity and $l_1 = 1$. The exponential function is the same as the Softmax function and we find these two functions have similar performances overall, so we only adopt the exponential function in the experiments.

\section{Implementation Details} \label{appendix:detail_implementation}
\subsection{Environment.} 
We conduct experiments under Python 3.8.5 and Pytorch 1.12.0. We use 4 Quadro RTX 8000 GPUs for computation.

\subsection{Data}
\textbf{Data partition.} To generate NonIID data partition amongst clients, we use Dirichlet distribution sampling in the trainset of each dataset. In our implementation, apart from clients having different class distributions, clients also have different dataset sizes; we think this partition is more realistic in practical scenarios. For the data partition in \autoref{fig:hyperplane_noniid} and \autoref{appdx_fig:toy_example}, we use a hybrid Dirichlet sampling to generate an FL system with both class-balanced clients and class-imbalanced clients. Specifically, we first generate all-client distribution with $\alpha_1$, and we only keep half of these clients. Then we use the remaining data to generate the distribution of remaining clients with $\alpha_2$. For the data in \autoref{fig:hyperplane_noniid}, we first generate a 20-client distribution with $\alpha_1 = 10$ and keep the first 10 clients as the balanced clients; then we use the remaining data to generate distributions of the last 10 imbalanced clients with $\alpha_2 = 0.1$. The distribution is shown in \autoref{appdx_fig:hyperplane_noniid_distr}.

\textbf{Data augmentation.} We adopt no data augmentation in the experiments. 

\textbf{Proxy dataset.} We use a small and class-balanced proxy dataset on the server. In \autoref{table:first_table}, we use proxy datasets with 10 samples per class, which means, for FashionMNIST and CIFAR-10, there are 100 samples in the proxy datasets, and for CIFAR-100, there are 1000 samples in the proxy datasets. The proxy datasets are randomly selected from the testset of each dataset. Then we use the remaining data in the testset to test the global models' performance for all compared methods. For \autoref{table:robustness} and the right of \autoref{fig:fedawo_partial}, we use CIFAR-10 and a 100-sample proxy dataset, while in \autoref{table:model_architetures}, we use CIFAR-10 and a 1000-sample proxy dataset.

\subsection{Model}
\textbf{SimpleCNN and MLP.} The SimpleCNN for CIFAR-10 and CIFAR-100 is a convolution neural network model with ReLU activations which consists of 3 convolutional layers followed by 2 fully connected layers. The first convolutional layer is of size (3, 32, 3) followed by a max pooling layer of size (2, 2). The second and third convolutional layers are of sizes (32, 64, 3) and (64, 64, 3), respectively. The last two connected layers are of sizes (64*4*4, 64) and (64, num\_classes, respectively. The MLP model for FasionMNIST is a three-layer MLP model with ReLU activations. The first layer is of size (28*28, 200), the second is of size (200, 200), and the last is (200, 10).

\textbf{ResNet and DenseNet.} We followed the model architectures used in \citep{li2018visualizing}. The numbers of the model names mean the number of layers of the models. Naturally, the larger number indicates a deeper network. For WRN56\_4 in \autoref{table:model_architetures}, it is an abbreviation of Wide-ResNet56-4, where "4" refers to four times as many filters per layer.

\subsection{Randomness}
Randomness is important for fair comparisons. In all experiments, we implement the experiments three times with different random seeds and report the averaged results. We use random seeds 8, 9, and 10 in all experiments. Given a random seed, we set torch, numpy, and random functions as the same random seed to make the data partitions and other settings identical. To make sure all algorithms have the same initial model, we save an initial model for each architecture and load the saved initial model at the beginning of one experiment. Also, for the experiments with partial participation, the participating clients in each round are vital in determining the model performance, and to guarantee fairness, we save the sequences of participating clients in each round and load the sequences in all experiments. This will make sure that, given a random seed and participation ratio, every algorithm will have the same sampled clients in each round.

\subsection{Evaluation}
We evaluate the global model performance on the testset of each dataset. The testset is mostly class-balanced and can reflect the global learning objective of an FL system. Therefore, we reckon the performance of the model on the testset can indicate the generalization performance of global models. In all experiments, we run 200 rounds and take the average test accuracy of the last 10 rounds as the final test accuracy for each experiment. For the indicators during training in \autoref{sect:gws}, like $\gamma$, $r$, the norm of global gradient, and the norm of GWS pseudo gradient, we take the averaged values in the middle stage of training, that is the average of 90-110 rounds.

\subsection{Hyperparameter}
\textbf{Learning rate and the scheduler.} We set the initial learning rates (LR) as 0.08 in CIFAR-10 and FashionMNIST and set LR as 0.01 in CIFAR-100. We set a decaying LR scheduler in all experiments; that is, in each round, the local LR is 0.99*(LR of the last round). 

\textbf{Local weight decay.} We adopt local weight decay in all experiments. For CIFAR-10 and FashionMNIST, we set the weight decay factor as 5e-4, and for CIFAR-100, we set it as 5e-5.

\textbf{Optimizer.} We set SGD optimizer as the clients' local solver and set momentum as 0.9. For the server-side optimizer (\textsc{FedDF, FedBE, Server-FT}, and \textsc{FedLAW}), we use Adam optimizer and betas=(0.5, 0.999).

\textbf{Hyperparameter for FL algorithms.} For \textsc{FedDF, FedBE} and \textsc{FedLAW}, we set the server epoch as 100. We observe for \textsc{Server-FT}, this epoch is too large that it will cause negative effects, so we set the epoch as 2 for \textsc{Server-FT}. We set $\mu_{FedProx} = 0.001$ in \textsc{FedProx} and $\alpha_{FedDyn} = 0.01$ in \textsc{FedDyn} as suggested in their official implementations or papers. For \textsc{FedBE}, we use the Gaussian mode in SWAG server. We did not use temperature smoothing in the ensemble distillation methods \textsc{FedDF} and \textsc{FedBE}.

\end{document}